\documentclass{IOS-Book-Article}

\usepackage{mathptmx}
\usepackage{soul}\setuldepth{article}
\usepackage{graphicx}
\usepackage{amsfonts}
\usepackage{amsmath}
\usepackage{amssymb}
\usepackage{xcolor}
\usepackage{blindtext}
\usepackage{color}
\usepackage{colortbl}
\usepackage{framed}
\usepackage{tcolorbox}

\def\hb{\hbox to 11.5 cm{}}

\begin{document}

\pagestyle{headings}
\def\thepage{}
\begin{frontmatter}              

\title{Responsible Artificial Intelligence: A Structured Literature Review}

\markboth{}{November 2023\hb}

\author[A]{\fnms{Sabrina} \snm{Göllner}
\thanks{Corresponding Author: Sabrina Göllner, sabrina.goellner@haw-hamburg.de}},
\author[A]{\fnms{Marina} \snm{Tropmann-Frick}
}
and
\author[B]{\fnms{Boštjan} \snm{Brumen}
}

\address[A]{Department of Computer Science, Hamburg University of Applied Sciences}
\address[B]{Faculty of Electrical Engineering and Computer Science, University of Maribor}

\begin{abstract}
Our research endeavors to advance the concept of responsible artificial intelligence (AI), a topic of increasing importance within EU policy discussions. The EU has recently issued several publications emphasizing the necessity of trust in AI, underscoring the dual nature of AI as both a beneficial tool and a potential weapon. This dichotomy highlights the urgent need for international regulation.
Concurrently, there's a need for frameworks that guide companies in AI development, ensuring compliance with such regulations. Our research aims to assist lawmakers and machine learning practitioners in navigating the evolving landscape of AI regulation, identifying focal areas for future attention.
This paper introduces a comprehensive and, to our knowledge, the first unified definition of responsible AI. Through a structured literature review, we elucidate the current understanding of responsible AI. Drawing from this analysis, we propose an approach for developing a future framework centered around this concept.
Our findings advocate for a human-centric approach to Responsible AI. This approach encompasses the implementation of AI methods with a strong emphasis on ethics, model explainability, and the pillars of privacy, security, and trust.
\end{abstract}

\end{frontmatter}

\section{Introduction}
In the past years, a lot of research is being conducted to improve Artificial Intelligence (AI) even further, as it is already being used in many aspects of life and industry.\\
The European Commision published a series of papers \cite{EU_WhitePaper.2020, EU_CoordinatedPlan.2021, EU_ProposalRegulation.2021} in which they address their strategy for AI.
In their white paper on AI from 2020 "A European Approach to Excellence and Trust“ the political options for promoting the use of AI while mitigating the risks associated with certain applications of this technology are set out.
This proposal aims to establish a legal framework for trustworthy AI in Europe so that the second objective of building an ecosystem for trust can be implemented.
The Framework should fully respect the values and rights of EU citizens.
It is repeatedly emphasized that AI should be human-centered and that European values have a high priority. The papers also address  challenging issues such as ethical issues, privacy, explainability, safety, and sustainability. It is pointed out how important security is in the context of AI and they also present a risk framework in five risk groups for AI systems in short form.
The document authors recognize that \textit{"[EU] Member States are pointing at the current absence of a common European framework."}
This indicates that a common EU framework is missing and it is an important political issue.

The document "Communication on Fostering a European Approach to AI“ represents a plan of the EU Commission, where numerous efforts are presented that are intended to advance AI in the EU or have already been undertaken.
In the beginning, it is stated that the EU wants to promote the development of a
\textit{"human-centric, sustainable, secure, inclusive and trustworthy artificial intelligence (AI) [which] depends on the ability of the European Union“}.

The Commission's goal is to ensure that excellence in the field of AI is promoted. Collaborations with stakeholders, building research capacity, environment for developers, and funding opportunities are talked about as well as bringing AI into the play for climate and environment. Part of the discussion on trust led to the question of how to create innovation. It was pointed out that the EU approach should be
\textit{"human-centered, risk-based, proportionate, and dynamic.“}\\
The plan also says they want to develop
\textit{"cutting-edge, ethical and secure AI, (and) promoting a human-centric approach in the global context“.}\\
At the end of the document there is an important statement:
\textit{"The revised plan, therefore, provides a valuable opportunity to strengthen competitiveness, the capacity for innovation, and the responsible use of AI in the EU“}.
The EC has also published the "Proposal for a Regulation laying down harmonized rules on artificial intelligence“ which contains, for example, a list of prohibited AI practices and specific regulations for AI systems that pose a high risk to health and safety as well as some transparency requirements.\\
It becomes noticeable that terms in the mentioned political documents that are used to describe the goal of trustworthy AI, however, keep changing (are inconsistent), and remain largely undefined.
The documents all reflect, on the one hand, the benefits and on the other hand the risks of AI from a political perspective.
It becomes clear that AI can improve our lives, solves problems in many ways, and is bringing added value but also can be a deadly weapon.
But on the other hand, the papers do not exactly define what trustworthy AI even means in concrete terms. Topics and subtopics are somehow addressed but there is no clear definition of (excellence and) trustworthiness, but more indirectly mentions some aspects which are important, e.g., ethical values, transparency, risks for safety as well as sustainability goals.\\
Furthermore, we believe that trust as a goal (as defined vaguely in the documents) is also not sufficient to deploy AI. Rather, we need approaches for a "responsible AI", which reflects on the EU values. This should of course also be trustworthy, but that concept covers just a part of the responsibility.
Therefore, in this paper, our goal is to find out the state-of-the-art from the scientific perspective and whether there is a general definition for "trustworthy AI". Furthermore, we want to clarify whether or not there is a definition for "responsible AI". The latter should actually be in the core of the political focus if we want to go towards \textit{"excellence“} in AI.

As a step towards responsible AI, we conduct a structured literature review that aims to provide a clear answer to what it means to develop a "responsible AI".

During our initial analysis, we found that there is a lot of inconsistency in the terminology overall, not only in the political texts. There is also a lot of overlap in the definitions and principles for responsible AI. In addition, similar/content-wise similar expressions exist that further complicate the understanding of responsible AI as a whole. There are already many approaches in the analyzed fields, namely trustworthy, ethical, explainable, privacy-preserving, and secure AI, but there are still many open problems that need to be addressed in the future.

Best to our knowledge this is one of the first detailed and structured reviews dealing with responsible AI.

The paper is structured as follows. In the following section, our research methodology is explained. This includes defining our research aims and objectives as well as specifying the databases and research queries we used for searching. The third section is the analysis part in which we first find out which definitions for responsible AI exist in the literature so far. Afterward, we explore content-wise similar expressions and look for their definitions in the literature. These are then compared with each other. As a result, we extract the essence of the analysis to formulate our definition of responsible AI. The subsequent section then summarizes the key findings in the previously defined scopes which are part of our definition of responsible AI.
We further conduct a qualitative analysis of every single paper regarding the terms "Trustworthy, Ethics, Explainability, Privacy, and Security" in a structured table and quantitative analysis of the study features. Furthermore, in the discussion part, we do specify the key points and describe the pillars for developing responsible AI. Finally, after mentioning the limitations of our work, we end with our conclusion and future work.


\section{Research Methodology}
To answer the research questions, a systematic literature review (SLR) was performed based on the guidelines developed in \cite{Kitchenham.2009}. The process of doing the structured literature review in our research is described in detail in the following subsections and summarized in the Systematic Review Protocol.

\subsection{Research Aims and Objectives}
In the present research, we aim to understand the role of "Responsible AI" from different perspectives, such as privacy, explainability, trust, and ethics. Firstly, our aim is to understand what constitutes the umbrella term "responsible AI", and secondly, to get an overview of the state of the art in the field. Finally, we seek to identify the open problems, challenges, and opportunities where further research is needed.\\

In summary, we provide the following contributions:
\begin{enumerate}
    \item Specify a concise Definition of "Responsible AI"
    \item Analyze the state of the art in the field of "Responsible AI"
\end{enumerate}

\subsection{Research Questions Formulation}
Based on the aims of the research, we state the following research questions:
\begin {itemize}
    \item RQ1: What is a general or agreed on definition of "Responsible AI" and what are the associated terms defining it?
    \item RQ2: What does "Responsible AI" encompass? 
\end {itemize}

\subsection{Databases}
In order to get the best results when searching for the relevant studies, we used the indexing data sources.
These sources enabled us a wide search of publications that would otherwise be overlooked. The following databases were searched:
\begin{itemize}
    \item ACM Digital Library (ACM)
    \item IEEE Explore (IEEE)
    \item SpringerLink (SL)
    \item Elsevier ScienceDirect (SD)
\end{itemize}
The reason for selecting these databases was to limit our search to peer-reviewed research papers only.

\subsection{Studies Selection}
To search for documents, the following search query was used in the different databases:\\
\texttt{("Artificial Intelligence" OR "Machine Learning" OR "Deep Learning" OR "Neural Network" OR "AI" OR "ML") AND (Ethic* OR Explain* OR Trust*) AND (Privacy*)}.\\
Considering that inconsistent terminology is used for "Artificial Intelligence", the terms "Machine Learning", "Deep Learning" and "Neural Network" were added, which should be considered synonyms.
Because there are already many papers using the abbreviations AI and ML, these were included to the set of synonyms.

The phrases "Ethic", "Trust" and "Explain" as well as "Privacy" was included with an asterisk (*), for all combinations of the terms following the asterisk, are included in the results (e.g. explain*ability).
The search strings were combined using the Boolean operator OR for inclusiveness and the operator AND for the intersection of all sets of search strings. These sets of search strings were put within parentheses.

The selection of the period of publication was set to two years: 2020 and 2021 to get all of the state-of-the-art papers. The search was performed in December 2021.

The results were sorted by relevance prior to the inspection, which was important because the lack of advanced options in some search engines returned many non-relevant results.

To exclude irrelevant papers, the authors followed a set of guidelines during the screening stage. Papers did not pass the screening if:

\begin{enumerate}
    \item They mention AI in the context of
    cyber-security, embedded systems, robotics, autonomous driving or internet of things, or alike.
    \item They are not related to the defined terms of responsible AI.
    \item They belong to general AI studies.
    \item They only consist of an abstract.
    \item They are published as posters.
\end{enumerate}

These defined guidelines were used to greatly decrease the number of full-text papers to be evaluated in subsequent stages, allowing the examiners to focus only on potentially relevant papers.

The initial search produced 10.313 papers of which 4.121 were retrieved from ACM, 1064 from IEEE, 1.487 from Elsevier Science Direct, and 3.641 from Springer Link. The screening using the title, abstract, and keywords removed 6.507 papers.
During the check of the remaining papers for eligibility, we excluded 77 irrelevant studies and 9 inaccessible papers.
We ended up with 254 papers that we included for the qualitative and quantitative analysis (see Figure 1).

\begin{figure}[h]
    \centering
    \includegraphics[height=8cm]{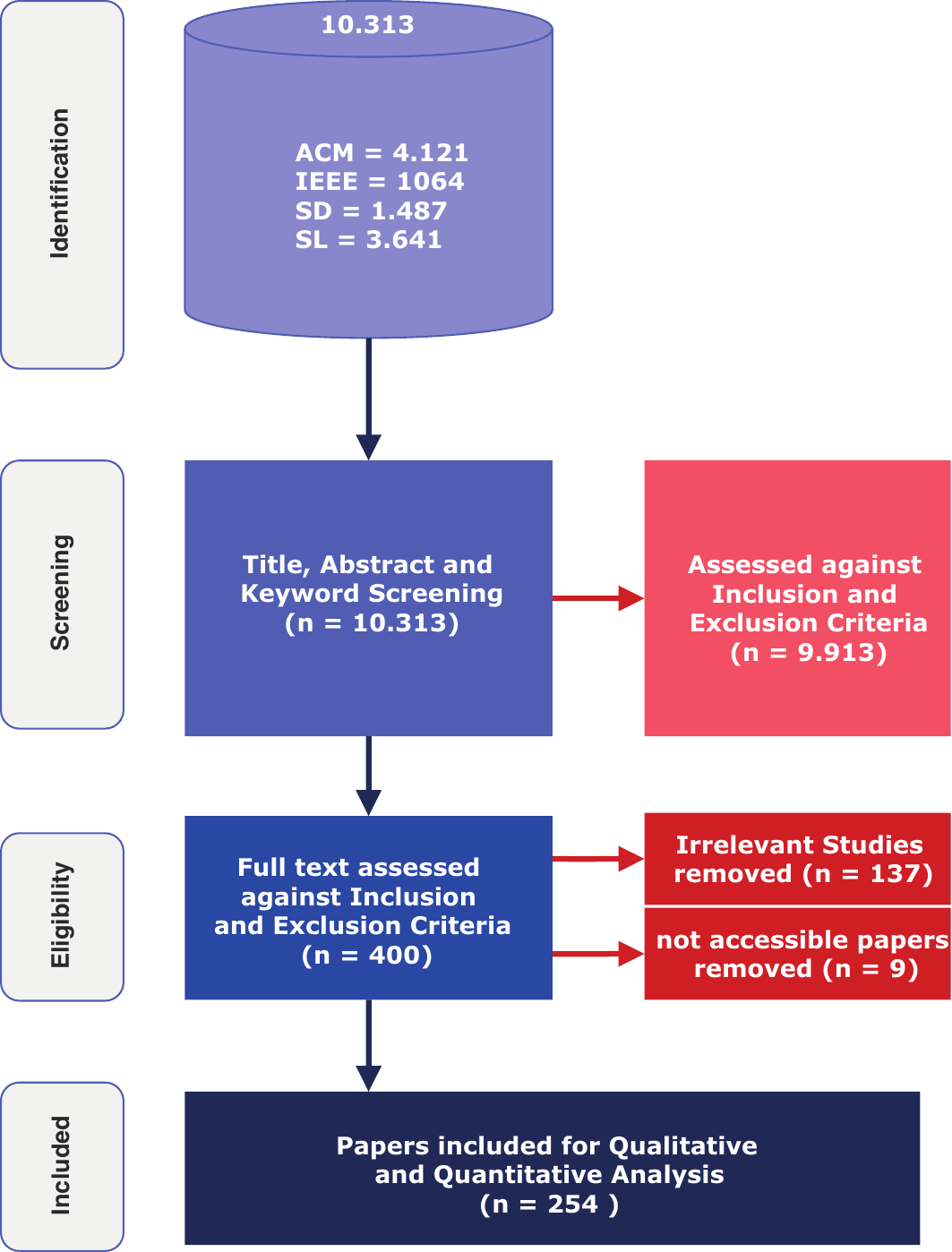}
    \caption{
        Structured review flow chart: the Preferred Reporting Items for Systematic Reviews and Meta-Analyses (PRISMA) flow chart detailing the records identified and screened, the number of full-text articles retrieved and assessed for eligibility, and the number of studies included in the review.
    }
    \label{fig:my_label}
\end{figure}

\section{Analysis}
\label{section:analysis}
This section includes the analysis part in which we first find out which definitions for 'responsible AI' existed in the literature so far. Afterward, we explore content-wise similar expressions and look for their definitions in the literature. These definitions are then compared with each other and searched for overlaps. As a result, we extract the essence of the analysis to formulate our definition of responsible AI.

\subsection{Responsible AI}

In this subsection, we answer the first research question: What is a general or agreed on definition of 'Responsible AI', and what are the associated terms defining it?

\subsubsection{Terms defining Responsible AI}
Out of all 254 analyzed papers, we only found 5 papers that explicitly introduce aspects for defining "responsible" AI.
The papers use the following terms in connection with 'responsible AI':

\begin{itemize}
    \item Fairness, Privacy, Accountability, Transparency and Soundness \cite{Maree.2020}
    \item Fairness, Privacy, Accountability, Transparency, Ethics, Security \& Safety \cite{AlejandroBarredoArrieta.2020}
    \item Fairness, Privacy, Accountability, Transparency, Explainability  \cite{EitelPorter.2021}
    \item Fairness, Accountability, Transparency, and Explainability \cite{werder_establishing_2022}
    \item Fairness, Privacy, Sustainability, Inclusiveness, Safety, Social Good, Dignity, Performance, Accountability, Transparency, Human Autonomy, Solidarity \cite{jakesch_how_2022}
\end{itemize}

However, after reading all 254 analyzed papers we strongly believe, that
the terms that are included in those definitions can be mostly treated as subterms or ambiguous terms.

\begin{itemize}
    \item 'Fairness'\cite{Maree.2020} and 'Accountability' \cite{Maree.2020,AlejandroBarredoArrieta.2020,EitelPorter.2021}, as well as the terms 'Inclusiveness, Sustainability, Social Good, Dignity, Human Autonomy, Solidarity' \cite{jakesch_how_2022} according to our definition, are subterms of Ethics.
    \item 'Soundness'\cite{Maree.2020}, interpreted as 'Reliability' or 'Stability', is included within Security and Safety.
    \item Transparency \cite{Maree.2020,AlejandroBarredoArrieta.2020,EitelPorter.2021} is often used as a synonym for explainability in the whole literature.
\end{itemize}

Therefore we summarize these terms of the above definitions to: "Ethics, Trustworthiness, Security, Privacy, and Explainability".
However, only the terms alone are not enough to get a picture of responsible AI.
Therefore, we will analyze and discuss what the \textit{meaning} of the five terms "Ethics, Trustworthiness, Security, Privacy, and Explainability" in the context of AI is, and how they \textit{depend} on each other.
During the analysis, we found also content-wise similar expressions to the concept of "responsible AI" which we want to include in the findings. This topic will be dealt with in the next section.

\subsubsection{Content-wise similar expressions for Responsible AI}

During the analysis, we found that
the term "Responsible AI" is often used interchangeably with the terms "Ethical AI" or "Trustworthy" AI, and "Human-Centered AI" is a content-wise similar expression.

Therefore, we treat the terms:
\begin{itemize}
    \item "Trustworthy AI", found in \cite{AI_HLEG.2019,Jain.2020,Sheth.2021,Wing.2021,Zhang.Qin.Li.2021,li_trustworthy_2022,strobel_data_2022}, and \cite{Kumar.2020} as cited in \cite{Floridi.2016}
    \item "Ethical AI", found in \cite{Hickok.2021,Loi.2020,Morley.2021,Ibanez.2021,Fjeld.2020}, and \cite{Milossi.2021} as cited in \cite{Floridi.2018}
    \item "Human-Centered AI", found in \cite{Shneiderman.2020} as cited in \cite{Fjeld.2020}
\end{itemize}
as the \textit{content-wise similar expressions} for "Responsible AI" hereinafter.

\subsection{Collection of definitions}
The resulting collection of definitions from 'responsible AI' and 'content-wise similar expressions for responsible AI' from the papers results in the following Venn diagram:

\begin{figure}[htb]
    \centering
    \begin{minipage}[c]{0.58\textwidth}
        \includegraphics[width=\textwidth]{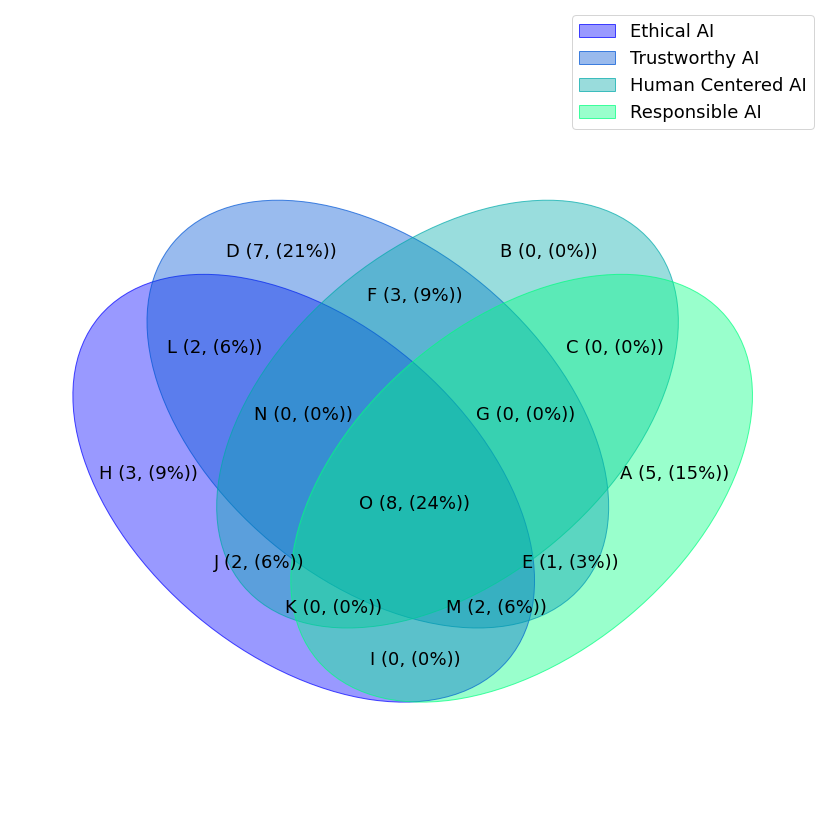}
    \end{minipage}
    \begin{minipage}[c]{0.38\textwidth}
        \centering
      \footnotesize{
        \begin{tabular}{|c|c|c|}
            \hline
            Set & Terms \\
            \hline
            \hline
            A & Solidarity, Performance, \\
                 & Sustainability, Soundness, \\
                 & Inclusiveness \\
            \hline
            B & - \\
            \hline
            C & - \\
            \hline
            D & Equality, Usability, \\
                 & Accuracy under Uncertainty, \\
                 & Assessment, Reliability, \\
                 & Data Control, Data Minimization \\
                 & Reproducibility, Generalization \\
                 & User Acceptance \\
            \hline
            E & Social Good \\
            \hline
            F & Human-Centered, Human Control, \\
            & Human Agency \\
            \hline
            G & - \\
            \hline
            H & Autonomy, Non-Maleficience, Trust \\
            \hline
            I & - \\
            \hline
            J & Human Values, Non-Discrimination \\
            \hline
            K & - \\
            \hline
            L & Compliant with Rules and Laws, \\
              & Social Robustness  \\
            \hline
            M & Human Autonomy, Dignity \\
            \hline
            N & - \\
            \hline
            O & Explainability, Safety, Fairness, \\
                 & Accountability, Ethics, Security \\
                 & Privacy, Transparency \\
            \hline

        \end{tabular}
    }
    \end{minipage}
    \caption{Venn diagram}%
  \end{figure}

 \paragraph{Analysis:}

We compared the definitions in the Venn diagram and determine the following findings:
\begin{itemize}
    \item From all four sets there is an overlap of 24\% of the terms: Explainability, Safety, Fairness, Accountability, Ethics, Security Privacy, Transparency.
    \item The terms occurring in the set of the definition for 'trust' only occurred in these, which is why this makes up the second largest set in the diagram. This is due to the fact that most of the terms actually come from definitions for trustworthy AI.
    \item There are also 6 null sets.
\end{itemize}

To tie in with the summary from the previous section, it should be pointed out once again that the terms 'Explainability, Safety, Fairness, Accountability, Ethics, Security Privacy, Transparency' can be grouped into generic terms as follows:
Ethics, Security, Privacy, and Explainability.

We also strongly claim that 'trust/trustworthiness' should be seen as an outcome of a responsible AI system, and therefore we determine, that it belongs to the set of requirements.
And each responsible AI should be built in a 'human-centered' manner, which makes it therefore another important subterm.

On top of these findings we specify our definition of Responsible AI in order to answer the first research question:


Responsible AI is \textcolor{blue}{\textbf{human-centered}} and ensures users' \textcolor{blue}{\textbf{trust}} through
\textcolor{blue}{\textbf{ethical}} ways of decision making.
The decision-making must be fair, accountable, not biased, with good intentions, non-discriminating, and consistent with societal laws and norms.
Responsible AI ensures, that automated decisions are  \textcolor{blue}{\textbf{explainable}} to users while always preserving users  \textcolor{blue}{\textbf{privacy}} through a \textcolor{blue}{\textbf{secure}} implementation.

As mentioned in the sections before, the terms defining "responsible AI" result from the analysis of the terms in sections 3.1.1 and 3.1.2. We presented a figure depicting the overlapping of the terms of content-wise similar expressions of Responsible AI, namely "Ethical AI, Trustworthy AI, and Human-Centered AI", and extracted the main terms of it. Also by summarizing the terms Fairness and Accountability into Ethics, and clarifying the synonyms (e.g., explainability instead of transparency), we finally redefined the terms defining "responsible AI" as \textbf{ "Human-centered, Trustworthy, Ethical, Explainable, Privacy(-preserving) and Secure AI"}.\\

\subsection{Aspects of Responsible AI}
\label{section:aspects_of_responsible_ai}
According to our analysis of the literature, we have identified several categories in section \ref{section:analysis} in connection to responsible AI, namely  "Human-centered, Trustworthy, Ethical, Explainable, Privacy-preserving and Secure AI" which should ensure the development and use of it. \\
To answer the second research question (RQ2), we analyze the state-of-the-art of topics "Trustworthy, Ethical, Explainable, Privacy-preserving and Secure AI" in the following subsections. We have decided to deal with the topic of 'Human-Centered AI' in a separate paper so as not to go beyond the scope of this work. \\

To find out the state of the art of the mentioned topics in AI, all 118 papers were assigned to one of the categories "Trustworthy AI, Ethical AI, Explainable AI, Privacy-preserving AI, and Secure AI", based on the prevailing content of the paper compared to each of the topic. These papers were then analyzed and we highlight their most important features of them in the following subsections.

\subsubsection{Trustworthy AI}

A concise statement for trust in AI is as follows:
\begin{quote}
    \textit{"Trust is an attitude that an agent will behave as expected and can be relied upon to reach its goal.
    Trust breaks down after an error or a misunderstanding between the agent and the trusting individual.
    The psychological state of trust in AI is an emergent property of a complex system, usually involving many cycles of design, training, deployment, measurement of performance, regulation, redesign, and retraining."}\cite{middleton_trust_2022}
\end{quote}

Trustworthy AI is about delivering the promise of AI's benefits while addressing the scenarios that have vital consequences for people and society. \\
In this subsection, we summarize which are the aspects covered by the papers in the category "Trustworthy AI" and what are the issues to engender users' trust in AI. \\

\paragraph{Surveys and Reviews}
 The following papers analyze trustworthy AI in their survey or review:
\cite{Jain.2020,Wing.2021,Zhang.Qin.Li.2021,Kumar.2020,Singh.2021,Beckert.2021}. The most important insights were the following:
\begin{itemize}
    \item According to \cite{Wing.2021} "\textit{Formal verification is a way to provide provable guarantees and thus increase one's trust that the system will behave as desired.}" However, this is more difficult with AI because of the inherently probabilistic nature of machine-learned models and the critical role of data in training, testing, and deploying a machine-learned model.
    \item The study of \cite{Beckert.2021} observes that implementation projects of Trustworthy AI from which best practices can be derived can only be found in the research contexts and not in the industry, with only a few exceptions. It is further suggested to break down existing implementation guidelines to the requirements of software engineers, computer scientists, and managers while embedding also social scientists or ethicists in the implementation process.
    \item The 'best practices' for Trusted AI formulated by \cite{Zhang.Qin.Li.2021} are Data and model transparency, data governance, data minimization, assessment methods (for fairness), and access requirements.
    \item The review of \cite{kaur_trustworthy_2023} has revolved around trustworthy AI and discusses its need and importance and requirements as well as testing techniques for verification.
\end{itemize}

\paragraph{Perception of trust}
The following publications deal with how humans perceive Trust in AI:
\cite{Araujo.2020,Knowles.2021,Lee.2021}. The interesting findings herein were as follows:
\begin{itemize}
    \item The study \cite{Araujo.2020} deals with analyzing users' trust in AI. Therefore, the authors examine the extent to which personal characteristics can be associated with perceptions of automated decision-making (ADM) through AI.
    The insight of the study was that Privacy can be seen as a central aspect as well as a human agency because people who felt they had more control over their own online information were more likely to view ADM as fair and useful.
    \item In the study of \cite{Knowles.2021} the authors found out that \textit{ "The general public are not users of AI; they are subject to AI."} There is a need for regulatory structures for trustworthiness.
    \item The study of \cite{Lee.2021} deals with Trust and Perceived Fairness around Healthcare AI and Cultural Mistrust. The key findings highlight that research around human experiences of AI should consider critical differences in social groups.
\end{itemize}

\paragraph{Frameworks}
Frameworks for "how developing Trustworthy AI can be achieved" were presented in \cite{Shneiderman.2020,Toreini.2020,Wang.2021}. The most important finding herein as the "chain of trust":
\begin{itemize}
    \item The authors in the study \cite{Toreini.2020} use various interrelated stages of a system life cycle within the development process for their concept. They then describe this process as forming the "Chain of Trust".
    \item \cite{nazaretsky_instrument_2022} introduce a "new instrument to measure teachers' trust in AI-based EdTech, provides evidence of its internal structure validity, and uses it to portray secondary-level school teachers' attitudes toward AI.“
    \item \cite{liao_designing_2022} develop a conceptual model called MATCH, which describes how trustworthiness is communicated in AI systems. They highlight transparency and interaction as AI systems' affordances that present a wide range of trustworthiness cues to users.
    \item \cite{seshia_toward_2022} consider the challenge of verified AI from the perspective of formal methods for making AI more trustworthy.
    \item \cite{li_trustworthy_2022}  we provide AI practitioners with a comprehensive guide for building trustworthy AI systems.
    \item \cite{banerjee_patient_2022} propose a framework and outlined case studies for applying modern data science to health care using a participatory design loop in which data scientists, clinicians, and patients work together.
    \item \cite{thuraisingham_trustworthy_2022} describes an architecture to support scalable trustworthy ML and describes the features that have to be incorporated into the ML techniques to ensure that they are trustworthy.
    \item \cite{choung_trust_2022} conceptualize trust in AI in a multidimensional, multilevel way and examine the relationship between trust and ethics.
\end{itemize}

\paragraph{Miscellaneous}
In other papers \cite{Jacovi.2021,PengHu.2021} related to "Trustworthy AI", we found the following:
\begin{itemize}
    \item Trust can be improved if the user can successfully understand the true reasoning process of the system (called intrinsic trust) \cite{Jacovi.2021}
    \item in the paper of \cite{holzinger_information_2022} is about information fusion as an integrative cross-sectional topic to gain more trustworthiness, robustness, and explainability.
    \item In the paper of \cite{gittens_adversarial_2022} reviews the state of the art in Trustworthy ML (TML) research and shed light on the multilateral tradeoffs, which are defined as the trade-offs among the four desiderata for TML they define as 'accuracy, robustness, fairness, and privacy' in the presence of adversarial attacks.
    \item \cite{allahabadi_assessing_nodate} showed how to assess Trustworthy AI (based on the EU guidelines) in practice in times of pandemic based on a deep-learning-based solution deployed at a public hospital.
    \item The article of \cite{strobel_data_2022} discusses the tradeoffs between data privacy and fairness, robustness as well as explainability in the scope of trustworthy ML.
    \item \cite{utomo_federated_2022} introduced the federated trustworthy Artificial Intelligence (FTAI) architecture.
\end{itemize}

Some papers we did not primarily categorize as "Trustworthy AI" (they rather belong to explainable AI) also mention important points dealing with trustworthiness:

\begin{itemize}
    \item According to \cite{Sheth.2021}, understanding AI is another important factor to achieve trust. Understanding means how AI-led decisions are made and what determining factors were included that are crucial to understanding. 
    \item Understanding is directly linked to the confidence if a model will act as intended when facing a given problem \cite{AlejandroBarredoArrieta.2020}.
    \item According to \cite{Burkart.2021}, in addition to understanding, also knowing the prediction model's strengths and weaknesses is important for gaining trust. 
\end{itemize}
We conclude that trust must be an essential goal of an AI application in order to be accepted in society and that every effort must be made to maintain and measure it at all times and in every stage of development. However, trustworthy AI still remains as a big challenge as it is not addressed (yet) holistically.

\subsubsection{Ethical AI}
In this subsection, we list the findings in the field of ethical AI. In our opinion, the definition found in \cite{Hanna.2021} best describes ethics in conjunction with AI:
\begin{quote}
    \textit{"AI ethics is the attempt to guide human conduct in the design and use of artificial automata or artificial machines, aka computers, in particular, by rationally formulating and following principles or rules that reflect our basic individual and social commitments and our leading ideals and values \cite{Hanna.2021}."}
\end{quote}

Now we come to summarize the most important key points that came up while analyzing the literature.

\paragraph{Reviews and Surveys}
\begin{itemize}
    \item \cite{stahl_european_2022} reviews the ethical and human rights challenges and proposed mitigation strategies to discuss how a regulatory body could be designed to address these challenges.
    \item \cite{huang_overview_2022} gives a comprehensive overview of the field of AI ethics, including a summary and analysis of AI ethical issues, ethical guidelines, and principles, approaches to addressing AI ethical issues, and methods for evaluating AI ethics.
    \item In the survey of \cite{petersen_responsible_2022} an overview of the technical and procedural challenges involved in creating medical machine learning systems responsibly and in conformity with existing regulations, as well as possible solutions to address these challenges, are discussed.
    \item \cite{benefo_ethical_2022} conducted a scientometric analysis of publications on the ethical, legal, social, and economic (ELSE) implications of artificial intelligence.
    \item \cite{karimian_ethical_2022} provide a systematic scoping review to identify the ethical issues of AI application in healthcare.
    \item \cite{attard-frost_ethics_nodate} conduct a semi-systematic literature review and thematic analysis to determine the extent to which the ethics of AI business practices are addressed in a wide range of guidelines.
    \item The review of \cite{tsamados_ethics_2022} contributes to the debate on the identification and analysis of the ethical implications of algorithms which aims to analyze epistemic and normative concerns and offer actionable guidance for the governance of the design, development, and deployment of algorithms.
\end{itemize}

\paragraph{Frameworks}
Implementing Ethical AI is often discussed and structured in frameworks because the difficulty in moving from principles to practice presents a significant challenge to the implementation of ethical guidelines. As also stated in \cite{Ibanez.2021}, there is still a significant gap. The following papers deal with solutions in the form of frameworks on this topic.
\begin{itemize}
    \item In \cite{Cheng.2021} the authors present a systematic framework for "socially responsible AI algorithms." The topics of AI indifference and the need to investigate socially responsible AI algorithms are addressed.
    \item \cite{Morley.2021} provided theoretical grounding of a concept named 'Ethics as a Service'.
    \item \cite{Benjamins.2021} developed a choices framework for the responsible use of AI for organizations which should help them to make better decisions toward the ethical use of AI. They distinguish between AI-specific technical choices e.g. continuous learning and generic digital technical choices, e.g., privacy, security, and safety.
    \item \cite{Bourgais.2021} proposed the "Ethics by design" framework which can be used to guide the development of AI systems. The "I" is based on three main aspects, "intelligibility, fairness, auditability" with the prototyping phase being crucial to establishing a solid ethical foundation for these systems.
    \item The article  \cite{EitelPorter.2021} also discusses about the possibility of developing ethical AI in a company by means of a framework. Different steps that lead through the whole development phase are discussed. It is also emphasized that rigorous testing and continuous measurement are of high importance to ensure that the system remains ethical and effective throughout its life cycle.
    \item In \cite{Peters.2020} two frameworks are presented including one for a responsible design process and another for better resolution of technology experience to help address the difficulty of moving from principle to practice in ethical impact assessment.
    \item \cite{VilleVakkuri.2021} presents "ECCOLA", a method using some kind of gamification method for implementing ethically aligned AI systems.
    \item \cite{contractor_behavioral_2022} advocates the use of licensing to enable legally enforceable behavioral use conditions on software and code and provides several case studies that demonstrate the feasibility of behavioral use licensing. It's envisioned how licensing may be implemented in accordance with existing responsible AI guidelines.
    \item \cite{joisten_focusing_2022} present TEDS as a new ethical concept, which focuses on the application of phenomenological methods to detect ethical errors in digital systems.
    \item \cite{bruschi_framework_2022} present a framework for assessing AI ethics and show applications in the field of cybersecurity.
    \item \cite{vyhmeister_responsible_2022} propose a framework for developing and designing AI components within the Manufacturing sector under responsible AI scrutiny (i.e. framework for developing ethics in/by design).
    \item \cite{belenguer_ai_2022} presents a novel approach for the assessment of the impact of bias to raise awareness of bias and its causes within an ethical framework of action.
    \item \cite{svetlova_ai_2022} relates the literature about AI ethics to the ethics of systemic risks, proposes a theoretical framework based on the ethics of complexity as well as applies this framework to discuss implications for AI ethics.
    \item \cite{li_fmea-ai_2022} propose an extension to FMEA, the "Failure mode and effects analysis", which is a popular safety engineering method, called “FMEA-AI” to support the conducting of “AI fairness impact assessments” in organizations.
    \item \cite{georgieva_ai_2022} are mapping AI ethical principles onto the lifecycle of an AI-based digital service and combining it with an explicit governance model to clarify responsibilities in operationalization.
    \item \cite{kumar_normative_2022} summarise normative ethical theories to a set of "principles for writing algorithms for the manufacture and marketing of artificially intelligent machines".
    \item \cite{solanki_operationalising_2022} offer a solution-based framework for operationalizing ethics in AI for healthcare.
    \item \cite{krijger_ai_2022} provide a holistic maturity framework in the form of an AI ethics maturity model that includes six critical dimensions for operationalizing AI ethics within an organization.
\end{itemize}

\paragraph{Tools}
\begin{itemize}
    \item \cite{wang_revise_2022} present a tool called: REvealing VIsual biaSEs (REVISE), that assists in the investigation of a visual dataset, surfacing potential biases along three dimensions: (1) object-based, (2) person-based, and (3) geography-based.
\end{itemize}

\paragraph{Ethical issues of AI}
The following papers discuss many of the existing ethical issues of AI:
\begin{itemize}
    \item \cite{Ayling.2021} identify the gaps in current AI ethics tools in auditing and risk assessment that should be considered.
    \item In \cite{Morley.2021} the \textit{explainability problem} deals with the fact that an AI black-box model is difficult to make understandable, and the \textit{public reason deficit}, the translation of code into a set of justifications in natural language.
    \item The work of \cite{Petrozzino.2021} looks more closely at the concept of ethical debt in AI and its consequences. The authors point out that the biggest challenge is seen here as the discrepancy between those who incur debt and those who ultimately pay for it. There is concern that the AI industry does little to address the complex sociotechnical challenges and that the industry is predominantly composed of individuals least likely to be affected by ethical debt.
    \item The contribution of \cite{Rochel.2020} to the literature on ethical AI concentrates on the work required to configure AI systems while addressing the AI engineer’s responsibility and refers to situations in which an AI engineer has to evaluate, decide and act in a specific way during the development.
    \item \cite{Stahl.2021} presents the findings of the "SHERPA project", which used case studies and a Delphi study to identify what people perceived to be ethical issues. The primary and frequent concern is privacy and data protection, which points to  more general questions about the reliability of AI systems. Lack of transparency makes it more difficult to recognize and address questions of bias and discrimination. Safety is also a key ethical issue; mostly this involves autonomous driving or systems for critical health services.
    It then addresses ethical issues arising from general artificial intelligence and sociotechnical systems that incorporate AI.
    \item \cite{Xiaoling.2021} points out different dilemmas like "Human Alienation", "Privacy Disclosure" and "Responsibility Issues".
    First, the author goes into various points such as human alienation (replacing human work with machines) which leads to higher unemployment rates; relying on smart technologies can lead to a decrease in independence; and the weakening of interpersonal relationships because of a closer relationship between man and machine. Second, the author addresses the issue of privacy leakage. He claims that service providers such as Google and Amazon are not complying with the General Data Protection Regulations in terms of completeness of information, clarity of language, and fairness of processing. Third, it will inevitably bring moral decision-making risks.
    \item \cite{Saetra.2021} shows up several ethical issues an AI-Ethicist should consider when making decisions and especially the dilemma when an AI ethicist must weigh the extent to which his or her own success in communicating a recognized problem involves a high risk of reducing the chances of successfully solving the problem. This is then resolved through different ethical theories (such as virtue ethics, deontological ethics, and consequentialist ethics).
    \item \cite{cooper_accountability_2022} reviewed Nissenbaum's "four barriers" to accountability, addressing the current situation in which data-driven algorithmic systems have become ubiquitous in decision contexts.
    \item The key finding of \cite{vakkuri_how_2022} is, that there is indeed a notable gap between the practices of the analyzed companies and the key requirements for ethical/trustworthy AI.
    \item \cite{weinberg_rethinking_2022} assesses and compares existing critiques of current fairness-enhancing technical interventions in machine learning that draw from a range of non-computing disciplines e.g. philosophy.
    \item \cite{lin_artificial_2022} explores the ethical issues of AI in environmental protection.
    \item The work of \cite{mulligan_ai_2022} outlines the ethical implications of AI from a climate perspective.
    \item The authors of \cite{waller_assembled_2022} make the case for the emergence of novel kinds of bias with the use of algorithmic decision-making systems.
    \item \cite{hagendorff_blind_2022} discusses blind spots regarding to topics that hold significant ethical importance but are hardly or not discussed at all in AI ethics.
    \item The critical discussion of \cite{bickley_cognitive_2022} argues for the application of cognitive architectures for ethical AI and
    \item \cite{fernandez-quilez_deep_2022} provide an overview of some of the ethical issues that both researchers and end users may face during data collection and development of AI systems, as well as an introduction to the current state of transparency, interpretability, and explainability of systems in radiological applications.
    \item \cite{munn_uselessness_2022} contributes critically to the ethical discussion of AI principles, arguing that they are useless because they cannot in any way mitigate the racial, social, and environmental harms of AI technologies, and seeks to suggest alternatives, thinking more broadly about systems of oppression and more narrowly about accuracy and auditing.
\end{itemize}

\paragraph{Miscellaneous}

The papers of \cite{Hagendorff.2020,Hickok.2021,Kiemde.2021,Zhou.2020b,Zhang.2021,Forbes.2021,Ibanez.2021,Morley.2021,Tartaglione.2020,Forsyth.2021,madaio_assessing_2022,tolmeijer_capable_2022} deal with ethical AI in their studies whereas they handled miscellaneous topics.

\begin{itemize}
    \item \cite{Hagendorff.2020,Forbes.2021} evaluates existing ethical frameworks.
    \item \cite{Hickok.2021}  gives a review of the documents that were published about ethical principles and guidelines and the lessons learned from them.
    \item \cite{Zhou.2020b}  surveyed the ethical principles and also their implementations. The paper suggested checklist-style questionnaires as benchmarks for the implementation of ethical principles of AI.
    \item \cite{Zhang.2021} collected insights from a survey of machine learning researchers.
    \item The study \cite{Forsyth.2021} reports the use of an interdisciplinary AI ethics program for high school students.
    Using short stories during the study was effective in raising awareness, focusing discussion, and helping students develop a more nuanced understanding of AI ethical issues, such as fairness, bias, and privacy.
    \item \cite{madaio_assessing_2022} provides an empirical study to investigate the discrepancies between the intended design of fairness mitigation tools and their practice and use in context. The focus is on: disaggregated assessments of AI systems designed to reveal performance differences across demographic groups.
    \item \cite{tolmeijer_capable_2022} compare AI and Human Expert Collaboration in Ethical Decision Making and investigate how the expert type (human vs. AI) and level of expert autonomy (adviser vs. decider) influence trust, perceived responsibility, and reliance.
    \item \cite{boyd_designing_2022} created a field guide for ethical mitigation strategies in machine learning through a web application.
    \item \cite{werder_establishing_2022} adds 'data provenance' as an important prerequisite to the table for mitigating biases stemming from the data's origins and pre-processing to realize responsible AI-based systems.
    \item \cite{chien_multi-disciplinary_2022} aims to provide a multi-disciplinary assessment of how fairness for machine learning fits into the context of clinical trials research and practice.
    \item \cite{lu_software_2022} perform an empirical study involving interviews with 21 scientists and engineers to understand the practitioners' views on AI ethics principles and their implementation.
    \item The work of \cite{nakao_toward_2022} explores the design of interpretable and interactive human-in-the-loop interfaces that enable ordinary end-users with no technical or domain knowledge background to identify and potentially address potential fairness issues.
    \item The article of \cite{rubeis_ihealth_2022} is an attempt to outline ethical aspects linked to iHealth by focussing on three crucial elements that have been defined in the literature: self-monitoring, ecological momentary assessment (EMA), and data mining.
    \item \cite{fabris_algorithmic_2022} have surveyed hundreds of datasets used in the fair ML and algorithmic equity literature to help the research community reduce its documentation debt, improve the utilization of existing datasets, and the curation of novel ones.
    \item \cite{belisle-pipon_artificial_nodate} surveyed the major ethical guidelines using content analysis and analyzed the accessible information regarding their methodology and stakeholder engagement.
    \item \cite{hausermann_community---loop_2022} introduce a business ethics perspective based on the normative theory of contractualism and conceptualize ethical implications as conflicts between the values of different interest groups.
    \item \cite{fung_confucius_2022} proposes a comparative analysis of the AI ethical guidelines endorsed by China and by the EU.
    \item The empirical study of \cite{starke_explainability_2022} deals with the ethics of using ML in psychiatric settings.
    \item \cite{stahl_computer_2022} compares the discourses of computer ethics with AI ethics and discusses their similarities, differences, issues, and social impact.
    \item The work of \cite{brusseau_ground_2022} is moving from the AI practice towards principles: Ethical insights are generated from the lived experiences of AI designers working on tangible human problems, and then cycled upward to influence theoretical debates.
    \item The main aim of \cite{anderson_ground_2022} has been to outline a new approach for AI ethics in heavy industry.
    \item \cite{ramanayake_immune_nodate} deals with research on pro-social rule breaking (PSRB) for AI.
    \item \cite{hunkenschroer_is_2022} aims to provide an ethical analysis of AI recruiting from a human rights perspective.
    \item \cite{valentine_recommender_2022} identify and discuss a set of advantages and ethical concerns related to incorporating recommender systems into the digital mental health ecosystem.
    \item The article of \cite{jacobs_reexamining_2022} focuses on the design and policy-oriented computer ethics while investigating new challenges and opportunities.
    \item The main goal of \cite{persson_future_2022} was to shed philosophical light on how the responsibility for guiding the development of AI in a desirable direction should be distributed between individuals and between individuals and other actors.
\end{itemize}

We also generally found during our analysis that Ethical AI deals often with fairness, therefore this should be mentioned here.
Fair AI can be understood as \textit{"AI systems [which] should not lead to any kind of discrimination against individuals or collectives in relation to race, religion, gender, sexual orientation, disability, ethnicity, origin or any other personal condition. Thus, fundamental criteria to consider while optimizing the results of an AI system is not only their outputs in terms of error optimization but also how the system deals with those groups."}\cite{AlejandroBarredoArrieta.2020}\\

In any case, the development of ethical artificial intelligence should be also subject to proper oversight within the framework of robust laws and regulations.\\

It is also stated, that transparency is widely considered also as one of the central AI ethical principles \cite{VilleVakkuri.2021}.

In the state-of-the-art overview of \cite{holzinger_towards_2022} the authors deal with the relations between explanation and AI fairness and examine, that fair decision-making requires extensive contextual understanding, and AI explanations help identify potential variables that are driving the unfair outcomes.

Mostly, transparency and explainability are achieved using so-called explainability (XAI) methods. Therefore, it is discussed separately in the following/next section.

\subsubsection{Explainable AI}
\label{section_explainble_ai}
Decisions made by AI systems or by humans using AI can have a direct impact on the well-being, rights, and opportunities of those affected by the decisions. This is what makes the problem of the explainability of AI such a significant ethical problem.
This subsection deals with the analysis of the literature in the field explainable AI (XAI).\\

We found an interesting definition in \cite{AlejandroBarredoArrieta.2020} which is quite suitable for defining explainable AI:
\begin{quote}
\textit{Given a certain audience, explainability refers to the details and reasons a model gives to make its functioning clear or easy to understand.\cite{AlejandroBarredoArrieta.2020}}
\end{quote}
In the following subsections, we highlight the most interesting aspects of XAI.

\paragraph{Black-box models problem}
According to \cite{AlejandroBarredoArrieta.2020} there is a trade-off between model explainability and performance. The higher accuracy comes at the cost of opacity: it is generally not possible to understand the reasons that explain why an AI system has decided the way it did, that it is the correct decision, or course of action was taken properly.
This is what, according to the literature, is often called interchangeably, AI's “black box,” “explainability,” “transparency,” “interpretability,” or “intelligibility” problem \cite{Maclure.2021} or "black box model syndrome" \cite{Vellido.2020}.

Another point to mention here is, that 'Explainable AI', which aims to open the black box of machine learning, might also be a Pandora's Box according to \cite{storey_explainable_2022}.
This means that opening the black box might undermine trust in an organization and its decision-making processes by revealing potential limitations of the data or model defects.

\cite{ratti_explainable_2022} claim also, that there is also a need for an explanation of how ML tools have been built, which requires documenting and justifying the technical choices that practitioners have made in designing such tools.

\paragraph{Synonyms for XAI}
These papers deal with the synonyms in context with XAI:
\begin{itemize}
    \item There is not yet consensus within the research community on the distinction between the terms interpretability, intelligibility, and explainability, and they are often, though not always, used interchangeably \cite{Kaur.2020, AlejandroBarredoArrieta.2020, Choras.2020}.
    \item \cite{Burkart.2021} says that usually, interpretability is used in the sense of understanding how the predictive model works as a whole. Explainability, on the other hand, is often used when explanations are given by predictive models that are themselves incomprehensible.
    \item \cite{GiuliaVilone.2021} mentioned 36 more notions related to the concept of explainability in their systematic review (e.g., 'Actionability', 'Causality', 'Completeness', 'Comprehensibility', 'Cognitive relief', etc.) They also provided a description of each of these notions.
    \item According to \cite{Brennen.2020} the lack of consistent terminology hinders the dialog about XAI.
\end{itemize}

\paragraph{Motivation for XAI}
The following papers address the motivation for XAI:
\begin{itemize}
    \item The key motivation of XAI is to \textit{"(1) increase the trustworthiness of the AI, (2) increase the trust of the user in a trustworthy AI, or (3) increase the distrust of the user in a non-trustworthy AI"} \cite{Jacovi.2021}.
    \item Explainability should be also considered as a bridge to avoid the unfair or unethical use of the algorithm's outputs.\cite{AlejandroBarredoArrieta.2020}
    \item According to \cite{Burkart.2021} other motivating aspects are causality, transferability, informativeness, fair and ethical decision-making, accountability, making adjustments, and proxy functionality.
    \item It should also help end-users to build a complete and correct mental model of the inferential process of either a learning algorithm or a knowledge-based system and to promote users' trust for its outputs, \cite{GiuliaVilone.2021} and reliance on the AI system \cite{Maltbie.2021}.
\end{itemize}

\paragraph{Reviews and Surveys}
\begin{itemize}
    \item \cite{rasheed_explainable_2022} have reviewed explainable and interpretable ML techniques for various healthcare applications while also highlighting security, safety, and robustness challenges along with ethical issues.
    \item \cite{saleem_explaining_2022} provides a survey, that attempts to provide a comprehensive review of global interpretation methods that completely explain the behavior of the AI models along with their strengths and weaknesses.
    \item \cite{tiddi_knowledge_2022} presents an extensive systematic literature review of the use of knowledge graphs in the context of Explainable Machine Learning.
    \item \cite{yang_unbox_2022} present a mini-review on explainable AI in health care, introducing solutions for XAI leveraging multi-modal and multi-center data fusion followed by two showcases of real clinical scenarios.
    \item \cite{saraswat_explainable_2022} proposed survey explicitly details the requirements of XAI in Healthcare 5.0, the operational and data collection process.
    \item The review of \cite{minh_explainable_2022} aims to provide a unified and comprehensive review of the latest XAI progress by discovering the critical perspectives of the rapidly growing body of research associated with XAI.
\end{itemize}

\paragraph{XAI Techniques}
There are many different XAI techniques discussed in the literature.
\cite{AlejandroBarredoArrieta.2020} as well as \cite{Burkart.2021} give a detailed overview of the known techniques and their strengths and weaknesses, therefore we will only cover this topic in short.\\
First, the models can be distinguished into two different approaches to XAI, the intrinsically transparent models and the Post-hoc explainability target models that are not readily interpretable by design. These so-called "black-box models" are the more problematic ones, because they are way more difficult to understand. The post-hoc explainability methods can then be distinguished further into model-specific and model-agnostic techniques.\\
We can also distinguish generally between data-independent and data-independent mechanisms for gaining interpretability as well as global and local interpretability methods.
\begin{itemize}
    \item \cite{zhang_debiased-cam_2022} highlight issues in explanation faithfulness when CNN models explain their predictions on images that are biased with systematic error and address this by developing Debiased-CAM to improve the truthfulness of explanations.
    \item In the work of \cite{golder_exploration_2022} a comprehensive analysis of the explainability of Neural Network models in the context of power Side-Channel Analysis (SCA) is presented, to gain insight into which features or Points of Interest (PoI) contribute to the most to the classification decision
    \item \cite{sun_investigating_2022} investigates the explainability of Generative AI for Code.
    \item In the work of \cite{patel_model_2022} provides a formal framework for achieving and analyzing differential privacy in model explanations and highlights the possible tradeoffs between fidelity of explanations and data privacy.
    \item \cite{terziyan_explainable_2022} deal with a transformation technique between black box models and explainable (as well as interoperable) classifiers on the basis of semantic rules via automatic recreation of the training datasets and retraining the decision trees (explainable models) in between.
    \item This research of \cite{rozanec_knowledge_2022} presented an architecture that supports the creation of semantically enhanced explanations for demand forecasting AI models.
    \item \cite{bacciu_explaining_2022} introduced LEGIT, a model-agnostic framework that incorporates the benefits of locally interpretable explanations into graph sampling methods.
    \item \cite{mery_black-box_2022} presents six different saliency maps that can be used to explain any face verification algorithm with no manipulation inside of the face recognition model.
    \item \cite{haffar_explaining_2022} focus on explaining by means of model surrogates the (mis)behavior of black-box models trained via federated learning.
\end{itemize}

\paragraph{Frameworks}
\cite{Sharma.2020} presents a single framework for analyzing the robustness, fairness, and explainability of a classifier based on counterfactual explanations through a genetic algorithm.

\paragraph{Application areas of explainability methods}
Application areas of explainability methods are, for example, medicine and health care, where these methods have a great influence. As discussed in \cite{Vellido.2020}, the authors paid specific attention to the methods of data and model visualization and concluded that involving the medical experts in the analytical process helped improve the interpretability and explainability of ML models even more.

\paragraph{Evaluation of explainability methods}
Not only the explainability methods but also the evaluation of those is of great relevance.
\begin{itemize}
    \item According to \cite{GiuliaVilone.2021} two main ways are objective evaluations and the other is human-centered evaluations. \item \cite{Sokol.2020} presented an explainability Fact Sheet Framework for guiding the development of new explainability approaches by aiding in their critical evaluation along the five proposed dimensions, namely: functional, operational, usability, safety, and validation.
    \item \cite{Mohseni.2021} presets a categorization of XAI design goals and evaluation methods. The authors used a mapping between design goals for different XAI user groups, namely: Novice Users, Data Experts, and AI Experts, and their evaluation methods. Further, they present a framework through a model and a series of guidelines to provide a high-level guideline for a multidisciplinary effort to build XAI systems.
    \cite{Jesus.2021} proposed the "XAI test", an application-based evaluation method tailored to isolate the effects of providing the end user with different levels of information. It became clear that the evaluation of XAI methods is still in the early stages and has to be very specific due to different end-user requirements.
    \item \cite{Maltbie.2021} analyzes XAI tools in the public sector. The case study based on a goal-question-metric analysis of explainability aims to quantitatively measure three state-of-the-art XAI tools. The results show that experts welcome new insights and more complex explanations with multiple causes. They also point out that the different levels of complexity may be appropriate for different stakeholders depending on their backgrounds.
    \item \cite{watson_agree_2022} show that the generated explanations are volatile when the model training changes, which is not consistent with the classification task and model structure. This raises further questions about confidence in deep learning models for healthcare.
    \item \cite{fel_how_2022} propose two new measures to evaluate explanations borrowed from the field of algorithmic stability: mean generalizability MeGe and relative consistency ReCo.
\end{itemize}

\paragraph{Stakeholders of XAI}
The target groups receiving the explanations need to be analyzed and their individual requirements are of great importance, as well as the usability of the software presenting these explanations:

\begin{itemize}
    \item XAI researchers often develop explanations based on their own intuition rather than the situated needs of their intended audience \cite{Ehsan.2021}.
    \item According to \cite{AlejandroBarredoArrieta.2020} it is very important that the generated explanations take into account the profile of the user who receives these explanations, the so-called audience.
    \item There are different user personalities to be considered, which probably require different explanation strategies, and these are not evenly covered by the current XAI tools \cite{Brennen.2020}.
    \item There are significant opportunities for UI/UX designers to contribute through new design patterns that make aspects of AI more accessible to different audiences \cite{Brennen.2020}.
    \item \cite{Burkart.2021} suggests building blocks into "What to explain" (content type), "How to explain" (communication), and "to Whom is the explanation addressed" (target group).
    \item According to the study of \cite{Sun.2021} explanation interfaces with User-Centric Explanation (“why”, “when”, and “how” different users need transparency information) as well as "Interactive Explanation" methods, which increases awareness of how AI agents make decisions, play an important role.
    \item \cite{Ehsan.2021} introduced Social Transparency (ST), a sociotechnically informed perspective that incorporates socio-organizational context in explaining AI-mediated decision-making. ST makes visible the socially situated technological context: the trajectory of AI decision outcomes to date, as well as people's interactions with those technological outcomes. Such contextual information could help people calibrate trust in AI, not only by tracking AI performance but also by incorporating human elements into AI that could elicit socially based perception and heuristics.
    \item In the study of \cite{Suresh.2021} proposes a three-tiered typology of stakeholder needs. It consists of long-term goals (understanding, building trust), shorter-term goals that work toward those goals (e.g., checking a model or questioning a decision), and immediate tasks that stakeholders can perform to achieve their goals (e.g., evaluating the reliability of predictions and detecting errors).
\end{itemize}

\paragraph{Miscellaneous}
\begin{itemize}
    \item \cite{tsiakas_using_2022} discuss the potential benefits of XAI and Human-In-The-Loop (HITL) methods in future work practices and illustrate how such methods can create new interactions and dynamics between human users and AI.
    \item The position paper of \cite{combi_manifesto_2022} brings together different roles and perspectives on XAI to explore the concept in-depth and offers a functional definition and framework for considering XAI in a medical context.
    \item \cite{hu_x-mir_2022} focus on the problem of explainable medical image retrieval using neural networks and different explanation methods.
    \item The article of \cite{padovan_black_2022} article seeks to provide technical explanations that can be given by XAI and to show how suitable explanations for liability can be reached in court.
\end{itemize}

The general public needs more transparency about how ML/AI systems can fail and what is at stake if they fail. Ideally, they should clearly communicate the outcomes and focus on the downsides to help people think about the trade-offs and risks of different choices (for example, the costs associated with different outcomes). But in addition to the general public also Data Scientists and ML Practitioners represent another key stakeholder group. In the study by \cite{Kaur.2020} the effectiveness and interpretability of two existing tools: the InterpretML implementation of GAMs and the SHAP Python package were investigated. Their results indicate that data scientists over-trust and misuse interpretability tools.\\
There is a “right to explanation” in the context of AI systems that directly affect individuals through their decisions, especially in legal and financial terms, which is one of the themes of the General Data Protection Regulation (GDPR) \cite{Choras.2020,Vellido.2020}.
Therefore we need to protect data through secure and privacy-preserving AI-methods. We will analyze this in the next section.

\subsubsection{Privacy-preserving and Secure AI}

As it was noted before, privacy and security are seen as central aspects of building trust in AI.
However, the fuel for the good performance of ML models is data, especially sensitive data. This has led to growing privacy concerns, such as unlawful use of private data and disclosure of sensitive data\cite{Cheng.2021, Abolfazlian.2020}. We, therefore, need comprehensive privacy protection through holistic approaches to privacy protection that can also take into account the specific use of data and the transactions and activities of users \cite{Bertino.2020} .\\
Privacy-preserving and Secure AI methods can help mitigate those risks. We define "Secure AI" as protecting data from malicious threats, which means protecting personal data from any unauthorized third-party access or malicious attacks and exploitation of data. It is set up to protect personal data using different methods and techniques to ensure data privacy. Data privacy is about using data responsibly. This means proper handling, processing, storage, and usage of personal information. It is all about the rights of individuals with respect to their personal information. Therefore data security is a prerequisite for data privacy.\\

\paragraph{Security and privacy threats}
There are a lot of security threats in the branch of machine learning like stealing the model or sensitive information from the user, reconstruction attacks, poisoning attacks, and membership inference attacks, while the latter is a rapidly evolving research branch \cite{Rahimian.2021}.
Selected papers deal with the security threats:
\begin{itemize}
    \item \cite{Ha.2020} provides a brief review of these threats as well as the defense methods on security and privacy issues in such models while maintaining their performance and accuracy. Therefore they classify three defense methods: gradient-level, function-level, and label-level, which are based on the differential privacy theory.
    \item In the paper of \cite{Rahimian.2021} several of these membership inference attacks across a large number of different datasets were evaluated as well as the Differential Private Stochastic Gradient Descent (DP-SGD) method as a defense.
    \item \cite{joos_adversarial_2022} contribute to this topic while providing an evaluation and critical reflection upon why prominent robustness methods fail to deliver a secure system despite living up to their promises of adding robustness in the light of facial authentication.
    \item \cite{jankovic_empirical_2022} provide an empirical Evaluation of Adversarial Examples Defences, Combinations, and Robustness Scores.
    \item \cite{brown_what_2022} argue, that a language model's privacy can be hardly preserved by such methods as for example Differential Privacy and conclude that the language model should be trained on text data that was explicitly produced for public use.
    \item \cite{giordano_adversarial_2022} study adversarial attacks on graph-level embedded methods.
\end{itemize}

The security threats need to be mitigated through techniques such as presented in the works of \cite{muhr_privacy-preserving_2022}, where a privacy-preserving detection of poisoning attacks in Federated Learning is presented and
in the paper of \cite{ma_shieldfl_2022}, which is about mitigating model poisoning in privacy-preserving Federated Learning.

\paragraph{Surveys and reviews on privacy-preserving techniques}
The following papers give an overview of privacy-preserving machine learning (PPML)-techniques through a survey:
\begin{itemize}
    \item \cite{AmineBoulemtafes.2020} evaluate privacy-preserving techniques in a comprehensive survey and propose a multi-level taxonomy, which categorizes the current state-of-the-art privacy-preserving deep learning techniques: (1) model training or learning, (2) PP inference or analysis, and (3) release a PP model.
    \item \cite{Chen.2020} summarize infrastructure support for privacy-preserving machine learning (PPML) techniques at both the software and hardware levels. The authors emphasize that the software/hardware co-design principle plays an important role in the development of a fully optimized PPML system.
    \item \cite{sousa_how_2022} provides a systematic review of deep learning methods for privacy-preserving natural language processing.
    \item \cite{zhang_visual_2022} have discussed visual privacy attacks and defenses in the context of deep learning in their survey.
\end{itemize}

The different PPML- techniques will be presented in the next few sections:
\paragraph{Differential Privacy}

Differential Privacy (DP) is a strict mathematical definition of privacy in the context of statistical and ML analyses \cite{Chen.2020} and many procedures are based mainly on this concept. "Differentially private" means to design query responses in such a way that it is impossible to detect the presence or absence of information about a particular individual in the database \cite{SergeyZapechnikov.2020}.
In the context of PPML, DP typically works by adding noise to the training database. The challenge is the trade-off between privacy and precision for a dataset. This means the amount of noise added to the data is what allows the quantification of privacy of the dataset \cite{Guevara.2021}.

\begin{itemize}
    \item DP approaches are described and used in \cite{Suriyakumar.2021,Zhu.2020,Harikumar.2021}.
    \item The study of \cite{Suriyakumar.2021} deals with the effects of differential privacy in the healthcare sector finding that DP-SDG is not well-suited for that kind of usage.
    \item The study of \cite{Harikumar.2021} deals with prescriptive analytics using DP using synthetic and real datasets and a new evaluation measure.
    \item The study of \cite{Zhu.2020} focuses on a practical method for private deep learning in computer vision based on a k-nearest neighbor.
    \item A a novel perturbed iterative gradient descent optimization (PIGDO) algorithm is proposed by \cite{ding_differentially_2021}.
    \item \cite{alishahi_add_2022} propose a novel Local Differential Privacy (LDP)-based feature selection system, called LDP-FS, that estimates the importance of features over securely protected data while protecting the confidentiality of individual data before it leaves the user's device.
    \item \cite{lal_deep_2022} used a deep privacy-preserving CTG data classification model by adopting the Differential Privacy (DP) framework.
    \item The major contribution of \cite{hassanpour_differential_2022} is adding differential privacy (DP) into continual learning (CL) procedures, aimed at protecting against adversarial examples.
    \item The paper of \cite{gupta_differential_2022} proposes a novel model based on differential privacy named DA-PMLM that protects the sensitive data and classification model outsourced by multiple owners in a real cloud environment.
    \item In \cite{liu_two-phase_2022} a new differential privacy decision tree building algorithm is proposed and secondly, this is used for developing a two-phase differential privacy random forest method which increases the complementarity among decision trees.
    \item \cite{zhao_correlated_2022} propose a novel correlated differential privacy of the multiparty data release (MPCRDP).
    \item \cite{arcolezi_differentially_2022} provides a comparative evaluation of differentially private DL models in both input and gradient perturbation settings for predicting multivariate aggregate mobility time series data.
\end{itemize}
DP is often used in combination with other techniques like Homomorphic Encryption, Federated Learning, and Secure Multiparty Computation.

\paragraph{Homomorphic Encryption}

Homomorphic Encryption (HE) allows performing computations directly with encrypted data (ciphertext) without the need to decrypt them. The method is typically used as follows: First, the owner of the data encrypts it using a homomorphic function and passes the result to a third party tasked with performing a specific calculation; the third party then performs the computation using the encrypted data and returns the result, which is encrypted because the input data is encrypted. The owner of the data then decrypts the result and receives the result of the calculation with the original plaintext data.
HE schemes support two types of computation: HE addition and HE multiplication \cite{Chen.2020}. Some noise is typically added to the input data during the encryption process. In order to get the expected result when decrypting, the noise must be kept below a certain threshold. This threshold affects the number of computations that can be performed on encrypted data.
The technique is used in the study of \cite{Yuan.2020}.
The following papers also deal with HE:
\begin{itemize}
    \item In the study of \cite{park_privacy-preserving_2022} a privacy-preserving training algorithm for a fair support vector machine classifier based on Homomorphic Encryption (HE) is proposed, where the privacy of both sensitive information and model secrecy can be preserved.
    \item \cite{liu_efficient_2022} propose a privacy-preserving logistic regression scheme based on CKKS, a leveled fully homomorphic encryption with the assistance of trusted hardware.
    \item \cite{byun_efficient_2022} proposed a privacy-preserving ridge regression algorithm with homomorphic encryption of multiple private variables and suggested an adversarial perturbation method that can defend attribute inference attacks on the private variables.
\end{itemize}

\paragraph{Secure Multiparty Computation}

Secure Multiparty Computation (MPC / SMPC) is a cryptographic protocol that distributes computation among multiple parties, where no single party can see the other parties' data. The parties are independent and do not trust each other. The main idea is to allow to perform computation on private data while keeping the data secret. MPC guarantees that all participants learn nothing more than what they can learn from the output and their own input.\\
In \cite{AnhTuTran.2021} this approach is used for a framework named "Secure Decentralized Training Framework" which is able to operate in a decentralized network that does not require a trusted third-party server while ensuring the privacy of local data with low communication bandwidth costs.\\
In the study of \cite{Wang.2020} the authors use this technique for the development of a novel Privacy-preserving Speech Recognition framework using the Bidirectional Long short-term memory neural network based on SMC.

\paragraph{Federated Learning}

Federated Learning (FL) is a popular framework for decentralized learning and FL is the most common method for preserving privacy found in this analysis.\\
The central idea is to have a base model first shared and then trained with each client node. The ML provider then creates a global model and sends it to the selected clients. The local models are then updated and improved via backpropagation using the local dataset. The global model is updated by aggregating the local updates (through federated averaging) only using the minimum necessary information \cite{Liu.2021}. FL ensures the privacy of the local participants since the client's data never leaves its local platform and no updates from individual users are stored in the cloud. Two different federated learning settings exist: cross-device (very large number of mobile or IoT devices) and cross-silo (a small number of clients) \cite{Liu.2021}.\\
In the state-of-the-art analysis of \cite{Yang.2021} the authors classify FL into different segmentations and algorithms used.
They also show current scenarios where FL is used including the Google GBoard System, Smart Medical Care, Smart Finance, Smart Transportation, and Smart Educational Systems.\\
Also, \cite{bonawitz_federated_2022} provides a brief introduction to key concepts in federated learning and analytics with an emphasis on how privacy technologies may be combined in real-world systems.
In the article of \cite{antunes_federated_2022} the authors specify their systematic literature on FL in the context of electronic health records for healthcare applications, whereas the survey of \cite{nguyen_federated_2023} is specifically about FL for smart healthcare.
In \cite{chowdhury2022review} the authors present an extensive literature review to identify state-of-the-art Federated Learning applications for cancer research and clinical oncology analysis.

The following papers use special FL-Approaches in their study:
\begin{itemize}
    \item In \cite{Diddee.2020} a user privacy preservation model for cross-silo Federated Learning systems (CrossPriv) is proposed.
    \item The study of \cite{Can.2021} a federated deep learning algorithm was developed for biomedical data collected from wearable IoT devices.
    \item In the literature FL is also used in \cite{Chen.2021} to create a federated parallel data platform (FPDP) including end-to-end data analytics pipeline.
    \item \cite{Fereidooni.2021} use FL for the design of SAFELearn, a generic private federated learning design that enables efficient thwarting of strong inference attacks that require access to clients' individual model updates.
    \item \cite{Hao.2021} presents an efficient, private and byzantine-robust FL (SecureFL) framework considering the communication and computation costs are reduced without sacrificing robustness and privacy protection.
    \item \cite{Li.2021} have been working on implementing SA for Python users in the context of the 'Flower FL Framework'.
    \item \cite{Xu.2021} introduce an approach for vertically partitioned FL setup achieving reduced training time and data-transfer time and enabling a changing sets of parties. (supports linear and regression models and SVM)
    \item \cite{Chai.2021} present FedAT, a novel Federated learning system with Asynchronous Tiers under Non-IID training data.
    \item \cite{Li_Hu.2021} proposing a scalable privacy-preserving federated learning (SPPFL) against poisoning attacks. The main contribution is crossing the chasm between these two contrary issues of poisoning defense and privacy protection.
    \item \cite{Cho.2021} introduces a new problem set in a multi-device context called Federated Learning in Multi-Device Local Networks (FL-MDLN) as well as highlighting the challenges of the proposed setting.
    \item \cite{Xu_Xhu.2021} presents a distributed FL framework in Trusted Execution Environment (TEE) to protect gradients from the perspective of hardware. The authors present the usage of trusted Software Guard eXtensions (SGX) as an instance to implement the FL as well as the proposal of an SGX-FL framework.
    \item The authors of \cite{Beilharz.2021} did not train a single global model, instead, the clients specialize in their local data in their approach and use other clients' model updates depending on the similarity of their respective data (based on a directed acyclic graph). The advantage of this approach are achieving more accuracy and less variance than through federated averaging.
    \item In this article of \cite{li_auto-weighted_2022}, the authors address the challenges of the standard FL techniques such as the vulnerability to data corruptions from outliers, systematic mislabeling, or even adversaries by proposing Auto-weighted Robust Federated Learning (ARFL), a novel approach that jointly learns the global model and the weights of local updates to provide robustness against corrupted data sources.
    \item The work of \cite{gong_cloudyfl_2022} deals with the challenge, that aggregating the data from different wearable devices to a central server introduces privacy concerns. Therefore they propose an architecture, CloudyFL, by deploying cloudlets close to wearable devices.
    \item \cite{kalloori_cross-silo_2022} designed a federated decision tree-based random forest algorithm using FL and conducted our experiments using different datasets. The test set was considering a small number of corporate companies for collaborative machine learning.
    \item \cite{zhu_fednkd_2022} propose FedNKD, which utilizes knowledge distillation and random noise, to enable federated learning to work dependably in the real world with complex data environments.
    \item \cite{wang_flare_2022} propose a robust model aggregation mechanism called FLARE for FL, which is designed for defending against state-of-the-art model poisoning attacks.
    \item In the paper of \cite{wang_blockchain_2022} a novel scheme based on blockchain architecture for Federated Learning data sharing is proposed.
    \item \cite{giuseppi_decentralized_2022} present a novel decentralized Federated Learning algorithm, DECFEDAVG, obtained as a direct decentralization of the original Federated Learning algorithm, FEDAVG.
    \item \cite{zhao_pvd-fl_2022} propose a privacy-preserving and verifiable decentralized federated learning framework, named PVD-FL.
    \item \cite{lo_towards_2022} present a blockchain-based trustworthy federated learning architecture to enhance the accountability and fairness of federated learning systems.
    \item In \cite{gholami_trusted_2022} proposed trust as a metric to enable secure federated learning through a mathematical framework for trust evaluation and propagation within a networked system.
    \item \cite{yang_trustworthy_2022} developed a Blockchain-FL architecture to ensure security and privacy, which utilizes secure global aggregation and blockchain techniques to resist attacks from malicious edge devices and servers.
    \item In \cite{abou_el_houda_when_2022} the authors design a new framework, called HealthFed, that leverages Federated Learning and blockchain technologies to enable privacy-preserving and distributed learning among multiple clinician collaborators.
    \item \cite{li_vfl-r_2022} propose "VFL-R", a novel Vertical FL framework combined with a ring architecture for multi-party cooperative modeling.
    \item \cite{sav_privacy-preserving_2022} presents a privacy-preserving federated learning-based approach, PriCell, for complex models such as convolutional neural networks.

\end{itemize}

Although Federated learning is a promising candidate for developing powerful models while preserving individual privacy and complying with the GDPR
the following papers show the challenges and vulnerabilities of FL:
\begin{itemize}
    \item The study of \cite{DivyaJatain.2021} highlights some vulnerabilities of this approach. Attacks in a federated setup can are classified as poisoning attacks (preventing the model to learn at all) or inference attacks (attacking the private data of the target participants).
    \item The study of \cite{Hao.2021} points out the weakness of destroying the integrity of the constructed model through byzantine attacks.
    \item The authors of \cite{DivyaJatain.2021} also point out that extensive communication is a big challenge too as well as system heterogeneity.
    \item Additionally to detecting these attacks and also identifying the attackers, \cite{Yang.2021} highlights the challenges of reducing communication overhead in the encryption process and solving the noise threshold of different scenarios. 
\end{itemize}

According to the survey of \cite{Yang.2021} an 'ideal state of FL' can be considered if the FL model can be fully decentralized and the current development makes it clear, that there are still many barriers on the way to this ideal state.\\
All in all, FL is still not enough to guarantee privacy, therefore it is often combined in hybrid mechanisms with other techniques as we will discuss in the section below.

\paragraph{Hybrid PPML-approaches}
\label{hybrid_priacy_approaches}
There are many new papers looking at hybridizing approaches as these could be promising solutions for the future. A hybrid PPML approach can take advantage of each component, providing an optimal tradeoff between ML task performance and privacy overhead\cite{Bertino.2020,AmineBoulemtafes.2020,Chen.2020}. The following papers deal with hybrid approaches:

\begin{itemize}
    \item The study of \cite{Chuanxin.2020} proposes an FL approach based on Gaussian differential privacy, called Noisy-FL, which can more accurately track the changes in privacy loss during model training.
    \item \cite{Jarin.2021} presents "PRICURE", a system that combines the complementary strengths of secure multiparty computation (SMPC) and differential privacy (DP) to enable privacy-compliant collaborative predictions between multiple model owners. SMPC is relevant for protecting data before inference, while DP targets post-inference protection to avoid attacks such as membership inference.
    \item The study of \cite{GrivetSebert.2021} proposes a deep learning framework building upon distributed differential privacy and a homomorphic argmax operator specifically designed to maintain low communication loads and efficiency.
    \item \cite{OwusuAgyemeng.2021} present a privacy-preserving DNN model known as Multi-Scheme Differential Privacy (MSDP) depending on the fusion of Secure Multi-party Computation (SMC) and $\epsilon$-differential privacy. The method reduces communication and computational cost at a minimal level.
    \item \cite{NuriaRodriguezBarroso.2020} presents the Sherpa.ai Federated Learning framework that is built upon a holistic view of federated learning and differential privacy. The study results both from exploring how the machine learning paradigm can be adapted to federated learning and from defining methodological guidelines for the development of artificial intelligence services based on federated learning and differential privacy.
    \item \cite{wibawa_homomorphic_2022} proposes a privacy-preserving federated learning algorithm for medical data using homomorphic encryption in a secure multi-party computation setting for protecting the DL model from adversaries.
    \item \cite{feng_data_2022} study the privacy protection strategy of enterprise information data based on consortium blockchain and federated learning.
\end{itemize}

\paragraph{Miscellaneous PPML-approaches}
There are a few interesting PPML approaches, as outlined below.
\begin{itemize}
    \item In \cite{Abuadbba.2020} the Split-learning method is used on 1D CNN models. Unfortunately, the results of the analysis show that it is possible to reconstruct the raw data from the activation of the split intermediate layer and this method needs further research.
    \item In \cite{Shayan.2021} a fully decentralized peer-to-peer (P2P) approach to multi-party ML, which uses blockchain and cryptographic primitives to coordinate a privacy-preserving ML process between peering clients is proposed which produces a final model that is similar in utility to federated learning and has the ability to withstand poisoning attacks.
    \item In \cite{Ghamry.2021} the authors propose a decentralized secure ML-training platform called Secular for based on using a private blockchain and InterPlanetary File System (IPFS) networks.
    \item In \cite{Agarwal.2020} the authors use an automatic face de-identification algorithm that generates a new face from a face image that retains the emotions and non-biometric facial attributes of a target face based on the StyleGAN technique.
    \item In \cite{Zhou.2020} the authors focus on designing an effective human-in-the-loop-aided (HitL-aided) scheme to preserve privacy in smart healthcare.
    \item \cite{zhou_human---loop-aided_2020} focuses on designing a human-in-the-loop-aided (HitL-aided) scheme to preserve privacy in smart healthcare.
    \item \cite{bai_privacy_2022} investigates and analyzes machine learning privacy risks to understand the relationship between training data properties and privacy leakage and  propose a privacy risk assessment scheme based on the clustering distance of training data.
    \item \cite{abbasi_privacy_2022} propose a comprehensive approach for face recognition techniques in a privacy preserving manner, i.e., without compromising the privacy of individuals in exchanged data while considering together the concepts of privacy and accuracy.
    \item \cite{montenegro_privacy-preserving_2022} contribute towards the development of more rigorous privacy-preserving methodologies capable of anonymizing case-based explanations without compromising their explanatory value.
    \item  \cite{mao_privacy-preserving_2022} designs a secure and efficient classification scheme based on SVM to protect the privacy of private data and support vectors in the calculation and transmission process.
    \item The authors of \cite{harichandana_privpas_2022} introduce PrivPAS (A real time Privacy-Preserving AI System) as a a novel framework to identify sensitive content.
    \item The authors of \cite{tian_sphinx_2022} present Sphinx, an efficient and privacy-preserving online deep learning system without any trusted third parties.
    \item The survey \cite{tan_towards_2022}  explores the domain of personalized FL (PFL) to address the fundamental challenges of FL on heterogeneous data, a universal characteristic inherent in all real-world datasets.
\end{itemize}

\paragraph{PPML-Measurement techniques}
Developing good techniques to preserve privacy is one thing but good techniques to measure it is needed as well.
In this regard, the paper by \cite{AmineBoulemtafes.2020} suggests these measurement techniques:
effectiveness, which is typically evaluated in terms of accuracy;
efficiency, which primarily includes communication or computational overhead and execution time; and
privacy, which is primarily evaluated in terms of direct and indirect guarantees against leakage.\\
We conclude that there is a lot of research related to privacy and security in the field of AI and there is no approach yet to achieve perfectly privacy-preserving and secure AI and many challenges are left open.

\subsection{Quantitative analysis}
The final set of 254 high-quality studies was selected for an in-depth analysis to aid in answering the presented research questions.

Our choice of features is based on their content in each of the following categories, "Trustworthy AI, Ethical AI, Explainable AI, Privacy-preserving AI, and Secure AI", as derived from section 3.2.
We analyzed the papers quantitatively.
Table \ref{table:quantitative_analysis} presents study features along with their absolute and percentile representations in the reviewed literature as well as their sources.\\

The distribution of the paper is as follows: most papers covered the topic "Privacy-Preserving and Secure AI", followed by "Ethical AI" and then "Explainable AI" and Trustworthy AI.\\
Within the topic "Privacy-Preserving and Secure AI", most papers belong to "Federated learning", obviously being a very emerging research field in the time frame.\\
There were also many different papers that were not assigned to any specific category (see "Miscellaneous)" since the topic is very multifaceted.\\
In the topic area of "Ethical AI", the most common category was 'Miscellaneous', since the authors of the ethical AI field handle very different topics.
In addition, second most of them could be assigned to the category 'ethical issues' since ths is a hot topic in the field of ethics.
The rest of the papers dealt with ethical frameworks that try to to integrate ethical AI in context of a development process.\\
Most studies in the field oxf XAI deal with coming up with new XAI approaches to solve different explainability problems with new AI models.
There were also a few that presented stakeholder analyses specifically in the context of explainability of AI models.
Few of them presented miscellaneous topics that could not be assigned to any specific category or frameworks to integrate explainable AI.\\
In Trustworthy AI, we saw that most presented a review or survey on the current state of Trustworthy AI in research.
There were also papers presented frameworks specially for trustwothiness or papers that reported on how Trust is perceived and described by different users.

\begin{center}
\begin{table}
    \footnotesize

    \rowcolors{2}{gray!25}{white}
    \tabcolsep=0.11cm

    \begin{tabular}{|c|c|c|c|}
    \hline
    \rowcolor{gray!50}
    Feature & Repr. & Perc. & Sources \\
    \hline
    \hline
    \rowcolor{gray!50}
    \multicolumn{4}{|c|}{Trustworthy AI (28/254, 11\% ) * } \\
    \hline
    \hline
    Reviews and Surveys & 9/28  &  32\%
    & \cite{Jain.2020,Kumar.2020,Singh.2021,Wing.2021,Beckert.2021,Zhang.Qin.Li.2021, kaur_trustworthy_2023,yang_unbox_2022,gittens_adversarial_2022}  \\
    \hline
    Perceptions of trust & 4/28  &  14\%
    & \cite{Araujo.2020,Knowles.2021,Lee.2021,middleton_trust_2022}\\
    \hline
    Frameworks & 9/28  &  32\%
    & \cite{Shneiderman.2020,Toreini.2020,Wang.2021,liao_designing_2022,seshia_toward_2022,li_trustworthy_2022,banerjee_patient_2022,thuraisingham_trustworthy_2022,choung_trust_2022}\\
    \hline
    Miscellaneous & 6/28  &  28\%
    & \cite{Jacovi.2021,PengHu.2021,holzinger_information_2022,allahabadi_assessing_nodate,strobel_data_2022,utomo_federated_2022}\\
    \hline
    \hline
    \rowcolor{gray!50}
    \multicolumn{4}{|c|}{Ethical AI ( 85/254,34\%) *} \\
    \hline
    \hline
    Frameworks & 19/85  &  22\%
    & \cite{Cheng.2021,Benjamins.2021,Bourgais.2021,EitelPorter.2021,Loi.2020,Peters.2020,VilleVakkuri.2021,Milossi.2021,contractor_behavioral_2022,joisten_focusing_2022} \\
    ~ & ~ & ~ &         \cite{bruschi_framework_2022,vyhmeister_responsible_2022,belenguer_ai_2022,svetlova_ai_2022,li_fmea-ai_2022,georgieva_ai_2022,kumar_normative_2022,solanki_operationalising_2022,krijger_ai_2022}\\
    \hline
    Ethical issues &  22/85  &  26\%
    & \cite{Ayling.2021,Loi.2020,Maclure.2021,Gambelin.2021,Xiaoling.2021,Gill.2021,Stahl.2021b,mulligan_ai_2022}\\
    ~ & ~ & ~ &          \cite{Rochel.2020,CharlesD.Raab.2020,Stahl.2021,Hanna.2021,Saetra.2021,Abolfazlian.2020,Petrozzino.2021,weinberg_rethinking_2022} \\
    ~ & ~ & ~ &          \cite{cooper_accountability_2022,vakkuri_how_2022,waller_assembled_2022,hagendorff_blind_2022,bickley_cognitive_2022,munn_uselessness_2022} \\
    \hline
    Miscellaneous & 33/85 & 39\%
    & \cite{Hagendorff.2020,Hickok.2021,Kiemde.2021,Zhou.2020b,Prunkl.2020,Zhang.2021,Forbes.2021,Ibanez.2021,Morley.2021,Tartaglione.2020,Forsyth.2021} \\
    ~ & ~ & ~ &
    \cite{madaio_assessing_2022,tolmeijer_capable_2022,boyd_designing_2022,chien_multi-disciplinary_2022,jakesch_how_2022,lu_software_2022,rubeis_ihealth_2022}\\
    ~ & ~ & ~ &
    \cite{valentine_recommender_2022,persson_future_2022,nakao_toward_2022,fabris_algorithmic_2022,belisle-pipon_artificial_nodate,hausermann_community---loop_2022,fung_confucius_2022} \\
    ~ & ~ & ~ &
    \cite{starke_explainability_2022,stahl_computer_2022,brusseau_ground_2022,anderson_ground_2022,ramanayake_immune_nodate,hunkenschroer_is_2022,jacobs_reexamining_2022,werder_establishing_2022}\\
    \hline
    Reviews and Surveys & 10/85 & 12\%
    & \cite{stahl_european_2022,rasheed_explainable_2022,huang_overview_2022,lin_artificial_2022,petersen_responsible_2022,benefo_ethical_2022,karimian_ethical_2022,attard-frost_ethics_nodate,tsamados_ethics_2022,holzinger_towards_2022} \\
    Tools & 1/85 & 1\% &
    \cite{wang_revise_2022}\\
    \hline
    \hline
    \rowcolor{gray!50}
    \multicolumn{4}{|c|}{Explainable AI ( 46/254 ,  18\%) *} \\
    \hline
    \hline
    Reviews and Surveys  & 10/46  &  22\% 
    & \cite{AlejandroBarredoArrieta.2020, Burkart.2021,Choras.2020,Sheth.2021,Sun.2021,Vellido.2020}\\
    ~ & ~ & ~ &
    \cite{GiuliaVilone.2021,saleem_explaining_2022,saraswat_explainable_2022,minh_explainable_2022}\\
    \hline
    Stakeholders & 7/46  &  15\% 
    & \cite{Brennen.2020,Ehsan.2021,Ehsan.2021b,Jesus.2021} \\
    ~ & ~ & ~ &
    \cite{Kaur.2020,Suresh.2021,Maltbie.2021}\\
    \hline
    XAI Approaches & 14/46  &  30\% 
    & \cite{AlexandreHeuillet.2021, Maree.2020, Sokol.2020, Yuan.2020, Zytek.2021, zhang_debiased-cam_2022}\\
    ~ & ~ & ~ &
    \cite{golder_exploration_2022,sun_investigating_2022,terziyan_explainable_2022,tiddi_knowledge_2022,bacciu_explaining_2022,mery_black-box_2022,haffar_explaining_2022,rozanec_knowledge_2022}\\
    \hline
    Frameworks & 4/46  &  9\% 
    & \cite{Mohseni.2021,Sharma.2020,Sokol.2020b, nazaretsky_instrument_2022}\\
    \hline
    Miscellaneous & 11/46  & 24\% 
    & \cite{Hailemariam.2020,Colaner.2021, patel_model_2022,tsiakas_using_2022,combi_manifesto_2022}\\
    ~ & ~ & ~ &
    \cite{watson_agree_2022,fel_how_2022,hu_x-mir_2022,padovan_black_2022,ratti_explainable_2022,storey_explainable_2022}\\
    \hline
    \hline

    \rowcolor{gray!50}
    \multicolumn{4}{|c|}{Privacy-preserving and Secure AI ( 95/254 ,  38\%) *} \\
    \hline
    \hline
    Reviews and Surveys  & 10/95  &  10\%
    & \cite{AmineBoulemtafes.2020,Chen.2020,Mercier.2021,Biswas.2021,Chang.2021,Bertino.2020} \\
    ~ & ~ & ~ &
    \cite{SergeyZapechnikov.2020,Liu.2021,sousa_how_2022,zhang_visual_2022}\\
    \hline
    Differential Privacy & 12/95  &  13\%
    & \cite{Harikumar.2021, Suriyakumar.2021, Zhu.2020, Guevara.2021, ding_differentially_2021,alishahi_add_2022} \\
    ~ & ~ & ~ &
    \cite{lal_deep_2022,hassanpour_differential_2022,gupta_differential_2022,liu_two-phase_2022,zhao_correlated_2022,arcolezi_differentially_2022}\\
    \hline
    Secure Multi-Party Computation & 2/95  & 2\%
    & \cite{AnhTuTran.2021,Wang.2020} \\
    \hline
    Homomorphic Encryption & 4/95  & 4\%
    & \cite{Yuan.2020,park_privacy-preserving_2022,liu_efficient_2022,byun_efficient_2022}\\
    \hline
    Federated learning &  35/95  &  37\%
    & \cite{Can.2021,Chen.2021,Diddee.2020,Fereidooni.2021,DivyaJatain.2021,ViraajiMothukuri.2021,Shayan.2021,Yang.2021} \\

    ~ & ~ & ~ &
    \cite{gong_cloudyfl_2022,kalloori_cross-silo_2022,zhao_pvd-fl_2022, Beilharz.2021,Hao.2021}   \\
    ~ & ~ & ~ &
    \cite{Li.2021,Xu.2021,Xu_Xhu.2021,Chai.2021,Cho.2021,Li_Hu.2021,Zhang.Yiu.Hui.2020,li_auto-weighted_2022,bonawitz_federated_2022}\\
    ~ & ~ & ~ &
    \cite{antunes_federated_2022,nguyen_federated_2023,zhu_fednkd_2022,wang_blockchain_2022,wang_flare_2022,giuseppi_decentralized_2022,lo_towards_2022}\\
    ~ & ~ & ~ &
    \cite{gholami_trusted_2022,yang_trustworthy_2022,abou_el_houda_when_2022,chowdhury2022review,sav_privacy-preserving_2022,ma_shieldfl_2022,li_vfl-r_2022}\\

    \hline

    Hybrid Approaches & 8/95  &  xx\%
    &
    \cite{Chuanxin.2020, GrivetSebert.2021, Jarin.2021, OwusuAgyemeng.2021, NuriaRodriguezBarroso.2020,wibawa_homomorphic_2022, feng_data_2022,tan_towards_2022}\\
    \hline
    Security Threats & 7/95 & 8\%
    &
    \cite{Rahimian.2021,Ha.2020,joos_adversarial_2022,jankovic_empirical_2022,brown_what_2022,muhr_privacy-preserving_2022,giordano_adversarial_2022}\\
    \hline
    Miscellaneous & 16/95 &  17\%
    & \cite{Agarwal.2020, Aminifar.2021, AnastasiiaGirka.2021, He.2020, Zhou.2020, Goldsteen.2021, Boenisch.2021, Abuadbba.2020, Ghamry.2021, zhou_human---loop-aided_2020} \\
    ~ & ~ & ~ &
    \cite{bai_privacy_2022,abbasi_privacy_2022,montenegro_privacy-preserving_2022,mao_privacy-preserving_2022,harichandana_privpas_2022,tian_sphinx_2022}\\
    \hline
\end{tabular}

\caption{Quantitative Analysis}
\label{table:quantitative_analysis}
*percentage does not add up to 100 due to rounding.

\end{table}
\end{center}

\clearpage

\subsection{Qualitative analysis}

The main categories of "Responsible AI", namely "Trustworthy AI, Ethical AI, Explainable AI, Privacy-preserving AI, and Secure AI", were defined in Section 3.2. The aspects of responsible AI were presented and discussed in detail in Section 3.3. Here, table 2 summarizes section 3.2. and 3.3. by presenting the qualitative analysis of the literature regarding the categories for responsible AI. Each of the papers was content-wise analyzed and checked for membership in the defined categories. \\

The legend in the table reads as follows: $\bullet$ = meets criteria (i.e., the focus of the paper covers the topic of the category; $\circ$ = partially meets criteria (i.e., the paper covers the topic of the category but its focus is elsewhere); no circle = does not meet criteria (i.e., the paper does deal with the topic of the category).
Abbreviations in the table heads are defined as follows: Trustworthy AI = Tr. AI, Ethical AI = Eth. AI, Explainable AI = XAI, Pivacy-preserving and Secure AI = PP \& Sec. AI \\

\begin{table}[h!]

\rowcolors{2}{gray!25}{white}
\begin{tabular}{|l|l|l|l|l|l|l|}
    \hline
    \rowcolor{gray!50}
     ~ & Author & Ref. & Tr. AI & Eth. AI & XAI & PP \& Sec. AI \\ \hline
    1  &  Abbasi et al. 2022 & \cite{abbasi_privacy_2022} & ~ & ~ & ~ & $\bullet$ \\ \hline
    2  &  Abou El Houda et al. 2022 & \cite{abou_el_houda_when_2022} & ~ & ~ & ~ & $\bullet$ \\ \hline
    3  &  Abolfazlian 2020 & \cite{Abolfazlian.2020} & $\bullet$ & $\bullet$ & $\circ$ & $\circ$ \\ \hline
    4  &  Abuadbba et al. 2020 & \cite{Abuadbba.2020} & ~ & ~ & ~ & $\bullet$ \\ \hline
    5  &  Agarwal et al. 2020 & \cite{Agarwal.2020} & ~ & ~ & ~ & $\bullet$ \\ \hline
    6  &  Allahabadi et al. 2022 & \cite{allahabadi_assessing_nodate} & $\bullet$ & $\circ$ & $\circ$ & $\circ$ \\ \hline
    7  &  Alishahi et al. 2022 & \cite{alishahi_add_2022} & ~ & ~ & ~ & $\bullet$ \\ \hline
    8  &  Anderson and Fort 2022 & \cite{anderson_ground_2022} & ~ & $\bullet$  & ~ & ~\\ \hline
    9  &  Antunes et al. 2022 & \cite{antunes_federated_2022} & ~ & ~ & ~ & $\bullet$ \\ \hline
    10 & Aminifar et al. 2021 & \cite{Aminifar.2021} & ~ & ~ & ~ & $\bullet$ \\ \hline
    11 & Araujo et al. 2020 & \cite{Araujo.2020} & $\bullet$ & $\bullet$ & ~ & $\circ$ \\ \hline
    12 & Arcolezzi et al. 2022 & \cite{arcolezi_differentially_2022} & ~ & ~ & ~ & $\bullet$ \\ \hline
    13 & Arrieta et al. 2020 & \cite{AlejandroBarredoArrieta.2020} & $\bullet$ & $\bullet$ & $\bullet$ & $\bullet$ \\ \hline
    14 & Attard-Frost et al. 2022 & \cite{attard-frost_ethics_nodate} & ~ & $\bullet$ & ~ & ~\\ \hline

\end{tabular}
\caption{Qualitative Analysis 1/8}
\label{able:qualitytive_analysis_1_8}
\end{table}

\begin{table}
    \rowcolors{2}{gray!25}{white}
    \begin{tabular}{|l|l|l|l|l|l|l|}
        \hline
        \rowcolor{gray!50}
        ~ & Author & Ref. & Tr. AI & Eth. AI & XAI & PP \& Sec. AI \\ \hline

        15 & Ayling and Chapman 2021 & \cite{Ayling.2021} & ~ & $\bullet$ & $\circ$ & $\circ$ \\ \hline
        16 & Bacciu and Numeroso 2022 & \cite{bacciu_explaining_2022} & ~ & ~ & $\bullet$ & ~ \\ \hline
        17 & Banerjee et al. 2022 & \cite{banerjee_patient_2022} & $\bullet$ & ~ & ~ & ~ \\ \hline
        18 & Bai et al. 2022 & \cite{bai_privacy_2022} & ~ & ~ & ~ & $\bullet$ \\ \hline
        19 & Beckert 2021 & \cite{Beckert.2021} & $\bullet$ & $\bullet$ & ~ & ~ \\ \hline
        20 & Belenguer 2022 & \cite{belenguer_ai_2022} & ~ & $\bullet$ & ~ & ~ \\ \hline
        21 & Bélisle-Pipon 2022 & \cite{belisle-pipon_artificial_nodate} & ~ & $\bullet$ & ~ & ~ \\ \hline
        22 & Beilharz et al. 2021 & \cite{Beilharz.2021} & ~ & ~ & ~ & $\bullet$ \\ \hline
        23 & Benefo et al. 2022 & \cite{benefo_ethical_2022} & ~ & $\bullet$ & ~ & ~\\ \hline
        24 & Benjamins 2021 & \cite{Benjamins.2021} & ~ & $\bullet$ & $\bullet$ & $\bullet$ \\ \hline
        25 & Bertino 2020 & \cite{Bertino.2020} & $\bullet$ & ~ & ~ & $\bullet$ \\ \hline
        26 & Bickley and Torgler 2021 & \cite{bickley_cognitive_2022} & ~ & $\bullet$ & ~ & ~ \\ \hline
        27 & Biswas 2021 & \cite{Biswas.2021} & ~ & ~ & ~ & $\bullet$ \\ \hline
        28 & Boenisch et al. 2021 & \cite{Boenisch.2021} & ~ & ~ & ~ & $\bullet$ \\ \hline
        29 & Bonawitz et al. 2022 & \cite{bonawitz_federated_2022} & ~ & ~ & ~ & $\bullet$ \\ \hline
        30 & Boulemtafes et al. 2020 & \cite{AmineBoulemtafes.2020} & ~ & ~ & ~ & $\bullet$ \\ \hline
        31 & Bourgais and Ibnouhsein 2021 & \cite{Bourgais.2021} & ~ & $\bullet$ & ~ & $\circ$ \\ \hline
        32 & Boyd 2022 & \cite{boyd_designing_2022} & ~ & $\bullet$ & ~ & ~ \\ \hline
        33 & Brennen 2020 & \cite{Brennen.2020} & $\circ$ & ~ & $\bullet$ & ~ \\ \hline
        34 & Brown et al. 2022 & \cite{brown_what_2022} & ~ & ~ & ~ & $\bullet$ \\ \hline
        35 & Brusseau 2022 & \cite{brusseau_ground_2022} & ~ & $\bullet$ & ~ & ~ \\ \hline
        36 & Bruschi and Diamede 2022 & \cite{bruschi_framework_2022} & ~ & $\bullet$ & ~ & ~ \\ \hline
        37 & Burkart and Huber 2021 & \cite{Burkart.2021} & $\circ$ & $\bullet$ & $\bullet$ & $\circ$ \\ \hline
        38 & Byun et al. 2022 & \cite{byun_efficient_2022} & ~& ~ & ~ & $\bullet$ \\ \hline
        39 & Can und Ersoy 2021 & \cite{Can.2021} & ~ & ~ & ~ & $\bullet$ \\ \hline
        40 & Chai et al. 2021 & \cite{Chai.2021} & ~ & ~ & ~ & $\bullet$ \\ \hline
        41 & Chang and Shokri 2021 & \cite{Chang.2021} & ~ & ~ & ~ & $\bullet$ \\ \hline
        42 & Chen et al. 2020 & \cite{Chen.2020} & ~ & ~ & ~ & $\bullet$ \\ \hline
        43 & Chen et al. 2021 & \cite{Chen.2021} & $\circ$ & ~ & ~ & $\bullet$ \\ \hline
        44 & Chien et al. 2022 & \cite{chien_multi-disciplinary_2022} & ~ & $\bullet$ & ~ & ~ \\ \hline
        45 & Cheng et al. 2021 & \cite{Cheng.2021} & $\bullet$ & $\bullet$ & $\bullet$ & $\bullet$ \\ \hline
        46 & Cho et al. 2021 & \cite{Chai.2021} & ~ & ~ & ~ & $\bullet$ \\ \hline
        47 & Choraś et al. 2020 & \cite{Choras.2020} & ~ & $\bullet$ & $\bullet$ & ~ \\ \hline
        48 & Choung et al. 2022 & \cite{choung_trust_2022} & ~ & $\bullet$ & ~ & ~ \\ \hline
        49 & Chowdhury et al. 2022 & \cite{chowdhury2022review} & ~ & ~ & ~ & $\bullet$ \\ \hline
    \end{tabular}
    \caption{Qualitative Analysis 2/8}
    \label{able:qualitytive_analysis_2_8}
\end{table}

\begin{table}
    \rowcolors{2}{gray!25}{white}
    \begin{tabular}{|l|l|l|l|l|l|l|}
        \hline
        \rowcolor{gray!50}
        ~ & Author & Ref. & Tr. AI & Eth. AI & XAI & PP \& Sec. AI \\ \hline
        50  & Chuanxin et al. 2020 & \cite{Chuanxin.2020} & ~ & ~ & ~ & $\bullet$ \\ \hline
        51  & Colaner et al. 2021 & \cite{Colaner.2021} & $\bullet$ & $\bullet$ & $\bullet$ & $\circ$ \\ \hline
        52 & Combi et al. 2022 & \cite{combi_manifesto_2022} & ~ & ~ & $\bullet$ & ~ \\ \hline
        53 & Contractor et al. 2022 & \cite{Colaner.2021} & ~ & $\bullet$ & ~ & ~ \\ \hline
        54 & Cooper et al. 2022 & \cite{cooper_accountability_2022} & ~ & $\bullet$ & ~ & ~ \\ \hline
        55  & Diddee and Kansra 2020 & \cite{Diddee.2020} & ~ & ~ & ~ & $\bullet$ \\ \hline
        56 & Ding et al. 2022 & \cite{ding_differentially_2021} & ~ & ~ & ~ & $\bullet$ \\ \hline
        57  & Ehsan et al. 2021 & \cite{Ehsan.2021} & $\bullet$ & ~ & $\bullet$ & ~ \\ \hline
        58  & Ehsan et al. 2021b & \cite{Ehsan.2021b} & $\bullet$ & ~ & $\bullet$ & ~ \\ \hline
        59  & Eitel-Porter 2021 & \cite{EitelPorter.2021} & $\circ$ & $\bullet$ & $\bullet$ & $\circ$ \\ \hline
        60 & Fabris et al. 2022 & \cite{fabris_algorithmic_2022} & ~ & $\bullet$ & ~ & ~ \\ \hline
        61 & Fel et al. 2022 & \cite{fel_how_2022} & ~ & $\bullet$ & ~ & ~ \\ \hline
        62  & Fereidooni et al. 2021 & \cite{Fereidooni.2021} & ~ & ~ & ~ & $\bullet$ \\ \hline
        63 & Fernandez-Quillez 2022 & \cite{fernandez-quilez_deep_2022} & ~ & $\bullet$ & ~ & ~ \\ \hline
        64 & Feng and Chen 2022 & \cite{feng_data_2022} & ~ & ~ & ~ & $\bullet$ \\ \hline
        65  & Forbes 2021 & \cite{Forbes.2021} & $\circ$ & $\bullet$ & ~ & ~ \\ \hline
        66  & Forsyth et al. 2021 & \cite{Forsyth.2021} & ~ & $\bullet$ & ~ & $\circ$ \\ \hline
        67 & Fung and Etienne 2022 & \cite{fung_confucius_2022} & ~ & $\bullet$ & ~ & ~ \\ \hline
        68  & Gambelin 2021 & \cite{Gambelin.2021} & ~ & $\bullet$ & ~ & ~ \\ \hline
        69  & Ghamry et al. 2021 & \cite{Ghamry.2021} & ~ & ~ & ~ & $\bullet$ \\ \hline
        70 & Gholami et al. 2022 & \cite{gholami_trusted_2022} & ~ & ~ & ~ & $\bullet$ \\ \hline
        71  & Gill 2021 & \cite{Gill.2021} & $\circ$ & $\bullet$ & $\circ$ & ~ \\ \hline
        72 & Giordano at al. 2022 & \cite{giordano_adversarial_2022} & ~ & ~ & ~ & $\bullet$ \\ \hline
        73  & Girka et al. 2021 & \cite{AnastasiiaGirka.2021} & ~ & ~ & ~ & $\bullet$ \\ \hline
        74 & Gittens et al. 2022 & \cite{gittens_adversarial_2022} & $\bullet$ & $\bullet$ & ~ & $\bullet$ \\ \hline
        75 & Giorgieva et al. 2022 & \cite{georgieva_ai_2022} & ~ & $\bullet$  & ~ & ~\\ \hline
        76 & Giuseppi et al. 2022 & \cite{giuseppi_decentralized_2022} & ~ & ~ & ~ & $\bullet$ \\ \hline
        77 & Golder et al. 2022 & \cite{golder_exploration_2022} & ~ & ~ & $\bullet$ & $\circ$ \\ \hline
        78  & Goldsteen et al. 2021 & \cite{Goldsteen.2021} & ~ & $\bullet$ & $\bullet$ & $\bullet$ \\ \hline
        79 & Gong et al. 2022 & \cite{gong_cloudyfl_2022} & ~ & ~ & ~ & $\bullet$ \\ \hline
        80  & Grivet Sébert et al. 2021 & \cite{GrivetSebert.2021} & ~ & ~ & ~ & $\bullet$ \\ \hline
        81  & Guevara et al. 2021 & \cite{Guevara.2021} & ~ & ~ & ~ & $\bullet$ \\ \hline
        82 & Gupta and Singh et al. 2022 & \cite{gupta_differential_2022} & ~ & ~ & ~ & $\bullet$ \\ \hline
        83  & Ha et al. 2020 & \cite{Ha.2020} & ~ & ~ & ~ & $\bullet$ \\ \hline
        84 & Haffar et al. 2022 & \cite{haffar_explaining_2022} & ~ & ~ & $\bullet$ & ~\\ \hline

    \end{tabular}
    \caption{Qualitative Analysis 3/8}
    \label{table:qualitytive_analysis_3_8}
\end{table}
\begin{table}
    \rowcolors{2}{gray!25}{white}
    \begin{tabular}{|l|l|l|l|l|l|l|}
        \hline
        \rowcolor{gray!50}
        ~ & Author & Ref. & Tr. AI & Eth. AI & XAI & PP \& Sec. AI \\ \hline
        85  & Hagendorff 2020 & \cite{Hagendorff.2020} & $\circ$ & $\bullet$ & $\bullet$ & $\bullet$ \\ \hline
        86 & Hagendorff 2022 & \cite{hagendorff_blind_2022} & ~ & $\bullet$ & ~ & ~ \\ \hline
        87  & Hailemariam et al. 2020 & \cite{Hailemariam.2020} & ~ & ~ & $\bullet$ & ~ \\ \hline
        88  & Hanna and Kazim 2021 & \cite{Hanna.2021} & ~ & $\bullet$ & ~ & $\circ$ \\ \hline
        89 & Häußermann and Lütge 2022 & \cite{hausermann_community---loop_2022} & ~ & $\bullet$ & ~ & ~ \\ \hline
        90  & Hao et al. 2021 & \cite{Hao.2021} & ~ & ~ & ~ & $\bullet$ \\ \hline
        91 & Harichandana et al. 2022 & \cite{harichandana_privpas_2022} & ~ & ~ & ~ & $\bullet$ \\ \hline
        92  & Harikumar et al. 2021 & \cite{Harikumar.2021} & $\bullet$ & ~ & ~ & $\bullet$ \\ \hline
        93 & Hassanpour et al. 2022 & \cite{hassanpour_differential_2022} & ~ & ~ & ~ & $\bullet$ \\ \hline
        94  & He et al. 2020 & \cite{He.2020} & ~ & ~ & ~ & $\bullet$ \\ \hline
        95  & Heuillet et al. 2021 & \cite{AlexandreHeuillet.2021} & $\circ$ & $\circ$ & $\bullet$ & ~ \\ \hline
        96  & Hickok 2021 & \cite{Hickok.2021} & ~ & $\bullet$ & ~ & ~ \\ \hline
        97 & Holzinger et al. 2022 & \cite{holzinger_information_2022} & $\circ$ & $\circ$ & $\circ$ & $\circ$ \\ \hline
        98  & Hu et al. 2021 & \cite{PengHu.2021} & $\bullet$ & ~ & ~ & ~ \\ \hline
        99 & Hu et al. 2022 & \cite{hu_x-mir_2022} & ~ & $\bullet$ & ~ & ~ \\ \hline
        100  & Huang et al. 2022 & \cite{huang_overview_2022} & ~ & $\bullet$ & ~ & ~ \\ \hline
        101   & Hunkenschroer \& Kriebitz 2022 & \cite{hunkenschroer_is_2022} & ~ & $\bullet$ & ~ & ~ \\ \hline
        102  & Ibáñez und Olmeda 2021 & \cite{Ibanez.2021} & $\circ$ & $\bullet$ & $\circ$ & $\circ$ \\ \hline
        103  & Jacovi et al. 2021 & \cite{Jacovi.2021} & $\bullet$ & $\circ$ & $\bullet$ & $\circ$ \\ \hline
        104 & Jacobs and Simon 2022 & \cite{jacobs_reexamining_2022} & ~ & $\bullet$ & ~ & ~ \\ \hline
        105 & Jakesch et al. 2022 & \cite{jakesch_how_2022} & ~ & $\bullet$ & ~ & ~ \\ \hline
        106  & Jain et al. 2020 & \cite{Jain.2020} & $\bullet$ & $\bullet$ & $\bullet$ & $\bullet$ \\ \hline
        107 & Jancovic \& Mayer 2022 & \cite{jankovic_empirical_2022} & ~ & ~ & ~ & $\bullet$ \\ \hline
        108  & Jarin and Eshete 2021 & \cite{Jarin.2021} & ~ & ~ & ~ & $\bullet$ \\ \hline
        109  & Jatain et al. 2021 & \cite{DivyaJatain.2021} & ~ & ~ & ~ & $\bullet$ \\ \hline
        110  & Jesus et al. 2021 & \cite{Jesus.2021} & $\circ$ & ~ & $\bullet$ & ~ \\ \hline
        111 & Joisten et al., 2022 & \cite{joisten_focusing_2022} & ~ & $\bullet$ & ~ & ~ \\ \hline
        112 & Joos et al., 2022 & \cite{joos_adversarial_2022} & ~ & ~ & ~ & $\bullet$ \\ \hline
        113 & Kalloori and Klingler 2022 & \cite{kalloori_cross-silo_2022} & ~ & ~ & ~ & $\bullet$ \\ \hline
        114 & Karimian et al. 2022 & \cite{karimian_ethical_2022} & ~ & $\bullet$ & ~ & ~ \\ \hline
        115  & Kaur et al. 2020 & \cite{Kaur.2020} & $\circ$ & ~ & $\bullet$ & ~ \\ \hline
        116 & Kaur et al. 2022 & \cite{kaur_trustworthy_2023} & $\bullet$ & ~ & ~ & ~ \\ \hline
        117  & Kiemde and Kora 2021 & \cite{Kiemde.2021} & ~ & $\bullet$ & ~ & ~ \\ \hline
        118  & Knowles and Richards 2021 & \cite{Knowles.2021} & $\bullet$ & $\circ$ & ~ & ~ \\ \hline
        119 & Krijger 2022 & \cite{krijger_ai_2022} & ~ & $\bullet$ & ~ & ~ \\ \hline

    \end{tabular}
    \caption{Qualitative Analysis 4/8}
    \label{table:qualitytive_analysis_4_8}
\end{table}
\begin{table}
    \rowcolors{2}{gray!25}{white}
    \begin{tabular}{|l|l|l|l|l|l|l|}
        \hline
        \rowcolor{gray!50}
        ~ & Author & Ref. & Tr. AI & Eth. AI & XAI & PP \& Sec. AI \\ \hline
        120  & Kumar et al. 2020 & \cite{Kumar.2020} & $\bullet$ & $\bullet$ & $\circ$ & $\circ$ \\ \hline
        121 & Kumar and Chowdhury 2022 & \cite{kumar_normative_2022} & ~ & $\bullet$ & ~ & ~\\ \hline
        122  & Lal and Kartikeyan 2022 & \cite{lal_deep_2022} & ~ & ~ & ~ & $\bullet$ \\ \hline
        123 & Lee and Rich 2021 & \cite{Lee.2021} & $\bullet$ & ~ & ~ & ~ \\ \hline
        124  & Li et al. 2021 & \cite{Li.2021} & ~ & ~ & ~ & $\bullet$ \\ \hline
        125  & Li et al. 2021 & \cite{Li_Hu.2021} & ~ & ~ & ~ & $\bullet$ \\ \hline
        126 & Li et al. 2022 & \cite{li_auto-weighted_2022} & ~ & ~ & ~ & $\bullet$ \\ \hline
        127 & Li et al. 2022 & \cite{li_trustworthy_2022} & $\bullet$ & ~ & ~ & ~ \\ \hline
        128 & Li et al. 2022 & \cite{li_fmea-ai_2022} & ~ & $\bullet$ & ~ & ~ \\ \hline
        129 & Li et al. 2022 & \cite{li_vfl-r_2022} & ~ & ~ & ~ & $\bullet$ \\ \hline
        130 & Liao and Sundar 2022 & \cite{liao_designing_2022} & $\bullet$ & ~ & ~ & ~ \\ \hline
        131 & Lin et al. 2022 & \cite{lin_artificial_2022} & ~ & $\bullet$ & ~ & ~ \\ \hline
        132  & Liu et al. 2021 & \cite{Liu.2021} & ~ & ~ & ~ & $\bullet$ \\ \hline
        133 & Liu et al. 2022 & \cite{liu_efficient_2022} & ~ & ~ & ~ & $\bullet$ \\ \hline
        134 & Liu et al. 2022 & \cite{liu_two-phase_2022} & ~ & ~ & ~ & $\bullet$ \\ \hline
        135 & Lo et al. 2022 & \cite{lo_towards_2022} & $\circ$ & $\circ$ & $\circ$ & $\bullet$ \\ \hline
        136  & Loi et al. 2020 & \cite{Loi.2020} & $\bullet$ & $\bullet$ & $\circ$ & ~ \\ \hline
        137 & Lu et al. 2022 & \cite{lu_software_2022} & $\circ$ & $\bullet$ & $\circ$ & $\circ$ \\ \hline
        138  & Maclure 2021 & \cite{Maclure.2021} & ~ & $\bullet$ & $\bullet$ & ~ \\ \hline
        139 & Madaio et al. 2022 & \cite{madaio_assessing_2022} & ~ & $\bullet$ & ~ & ~ \\ \hline
        140  & Maltbie et al. 2021 & \cite{Maltbie.2021} & $\circ$ & ~ & $\bullet$ & ~ \\ \hline
        141 & Mao et al. 2022 & \cite{mao_privacy-preserving_2022} & ~ & ~ & ~ & $\bullet$ \\ \hline
        142  & Ma et al. 2022 & \cite{ma_shieldfl_2022} & ~ & ~ & ~ & $\bullet$ \\ \hline
        143  & Maree et al. 2020 & \cite{Maree.2020} & ~ & ~ & $\bullet$ & ~ \\ \hline
        144  & Mercier et al. 2021 & \cite{Mercier.2021} & ~ & ~ & ~ & $\bullet$ \\ \hline
        145 & Mery and Morris 2022 & \cite{mery_black-box_2022} & ~ & ~ & $\bullet$ & ~ \\ \hline
        146 & Middleton et al. 2022 & \cite{middleton_trust_2022} & $\bullet$ & ~ & ~ & ~ \\ \hline
        147  & Milossi et al. 2021 & \cite{Milossi.2021} & $\circ$ & $\bullet$ & ~ & $\circ$ \\ \hline
        148 & Minh et al. 2021 & \cite{minh_explainable_2022} & ~ & ~ & $\bullet$ & ~ \\ \hline
        149  & Mohseni et al. 2021 & \cite{Mohseni.2021} & $\bullet$ & ~ & $\bullet$ & ~ \\ \hline
        150  & Montenegro et al. 2022 & \cite{montenegro_privacy-preserving_2022} & ~ & ~ & ~ & $\bullet$ \\ \hline
        151  & Morley et al. 2021 & \cite{Morley.2021} & ~ & $\bullet$ & $\circ$ & $\circ$ \\ \hline
        152  & Mothukuri et al. 2021 & \cite{ViraajiMothukuri.2021} & ~ & ~ & ~ & $\bullet$ \\ \hline
        153 & Muhr et al. 2021 & \cite{muhr_privacy-preserving_2022} & ~ & ~ & ~ & $\bullet$ \\ \hline
        154 & Mulligan \& Elaluf-Calderwood 2022 & \cite{mulligan_ai_2022} & ~ & $\bullet$ & ~ & ~ \\ \hline
\end{tabular}
\caption{Qualitative Analysis 5/8}
\label{table:qualitytive_analysis_5_8}
\end{table}

\begin{table}
    \rowcolors{2}{gray!25}{white}
    \begin{tabular}{|l|l|l|l|l|l|l|}
        \hline
        \rowcolor{gray!50}
        ~ & Author & Ref. & Tr. AI & Eth. AI & XAI & PP \& Sec. AI \\ \hline

        155 & Munn 2022 & \cite{munn_uselessness_2022} & ~ & $\bullet$ & ~ & ~ \\ \hline
        156 & Nakao et al., 2022 & \cite{nakao_toward_2022} & ~ & $\bullet$ & ~ & ~ \\ \hline
        157 & Nazaretsky et al., 2022 & \cite{nazaretsky_instrument_2022} & $\bullet$ & ~ & ~ & ~ \\ \hline
        158 & Nguyen et al., 2022 & \cite{nguyen_federated_2023} & ~ & ~ & ~ & $\bullet$ \\ \hline
        159  & Owusu-Agyemeng et al. 2021 & \cite{OwusuAgyemeng.2021} & ~ & ~ & ~ & $\bullet$ \\ \hline
        160 & Padovan et al. 2022 & \cite{padovan_black_2022} & ~ & ~ & $\bullet$ & ~ \\ \hline
        161 & Park et al. 2022 & \cite{park_privacy-preserving_2022} & ~ & $\circ$ & ~ & $\bullet$ \\ \hline
        162 & Patel et al. 2022 & \cite{patel_model_2022} & ~ & ~ & $\bullet$ & $\bullet$ \\ \hline
        163 & Persson \& Hedlund 2022 & \cite{persson_future_2022} & ~ & $\bullet$ & ~ & ~ \\ \hline
        164  & Peters et al. 2020 & \cite{Peters.2020} & ~ & $\bullet$ & $\bullet$ & $\circ$ \\ \hline
        165 & Petersen et al. 2022 & \cite{petersen_responsible_2022} & $\circ$ & $\circ$& $\circ$ & $\circ$ \\ \hline
        166  & Petrozzino 2021 & \cite{Petrozzino.2021} & ~ & $\bullet$ & ~ & ~ \\ \hline
        167  & Prunkl and Whittlestone 2020 & \cite{Prunkl.2020} & ~ & $\bullet$ & ~ & ~ \\ \hline
        168  & Raab 2020 & \cite{CharlesD.Raab.2020} & $\bullet$ & $\bullet$ & $\circ$ & $\bullet$ \\ \hline
        169  & Rahimian et al. 2021 & \cite{Rahimian.2021} & ~ & ~ & $\bullet$ & $\bullet$ \\ \hline
        170 & Ramanayake et al. 2021 & \cite{ramanayake_immune_nodate} & ~ & $\bullet$ & ~ & ~ \\ \hline
        171   & Rasheed et al. 2022 & \cite{rasheed_explainable_2022} & $\circ$ & $\circ$ & $\bullet$ & $\circ$ \\ \hline
        172 & Ratti et al. 2022 & \cite{ratti_explainable_2022} & ~ & ~ & $\bullet$ & ~ \\ \hline
        173  & Rochel and Evéquoz 2020 & \cite{Rochel.2020} & ~ & $\bullet$ & ~ & ~ \\ \hline
        174  & Rodríguez-Barroso et al. 2020 & \cite{NuriaRodriguezBarroso.2020} & ~ & ~ & ~ & $\bullet$ \\ \hline
        175  & Rozanec et al. 2022 & \cite{rozanec_knowledge_2022} & ~ & ~ & $\bullet$ & ~ \\ \hline
        176 & Rubeis 2022 & \cite{rubeis_ihealth_2022} & ~ & $\bullet$ & ~ & ~ \\ \hline
        177  & Saetra et al. 2021 & \cite{Saetra.2021} & $\bullet$ & ~ & ~ & ~ \\ \hline
        178  & Saleem et al. 2022 & \cite{saleem_explaining_2022} & ~ & ~ & ~ & $\bullet$ \\ \hline
        179 & Saraswat et al. 2022 & \cite{saraswat_explainable_2022} & ~ & ~ & $\bullet$ & ~ \\ \hline
        180 & Sav et al. 2022 & \cite{sav_privacy-preserving_2022} & ~ & ~ & ~ & $\bullet$ \\ \hline
        181  & Sharma et al. 2020 & \cite{Sharma.2020} & ~ & $\bullet$ & $\bullet$ & ~ \\ \hline
        182  & Shayan et al. 2021 & \cite{Shayan.2021} & ~ & ~ & ~ & $\bullet$ \\ \hline
        183 & Seshia et al. 2022 & \cite{seshia_toward_2022} & $\bullet$ & ~ & ~ & $\circ$ \\ \hline
        184  & Sheth et al. 2021 & \cite{Sheth.2021} & $\bullet$ & ~ & $\bullet$ & ~ \\ \hline
        185  & Shneiderman et al. 2020 & \cite{Shneiderman.2020} & $\bullet$ & $\bullet$ & $\circ$ & $\circ$ \\ \hline
        186  & Singh et al. 2021 & \cite{Singh.2021} & $\bullet$ & $\bullet$ & $\bullet$ & $\bullet$ \\ \hline
        187  & Sokol and Flach 2020 & \cite{Sokol.2020} & ~ & ~ & $\bullet$ & $\circ$ \\ \hline
        188  & Sokol and Flach 2020b & \cite{Sokol.2020b} & $\bullet$ & ~ & $\bullet$ & ~ \\ \hline

\end{tabular}
\caption{Qualitative Analysis 6/8}
\label{table:qualitytive_analysis_6_8}
\end{table}

\begin{table}
    \rowcolors{2}{gray!25}{white}
    \begin{tabular}{|l|l|l|l|l|l|l|}
        \hline
        \rowcolor{gray!50}
        ~ & Author & Ref. & Tr. AI & Eth. AI & XAI & PP \& Sec. AI \\ \hline

        189 & Solanki 2022 & \cite{solanki_operationalising_2022} & ~ & $\bullet$ & ~ & ~ \\ \hline
        190 & Sousa and Kern & \cite{sousa_how_2022} & ~ & ~ & ~ & $\bullet$ \\ \hline
        191  & Stahl 2021 & \cite{Stahl.2021} & ~ & $\bullet$ & $\bullet$ & $\bullet$ \\ \hline
        192  & Stahl et al. 2021 & \cite{Stahl.2021b} & $\bullet$ & $\bullet$ & ~ & $\circ$ \\ \hline
        193 & Stahl et al. 2022 & \cite{stahl_european_2022} & ~ & $\bullet$ & ~ & ~ \\ \hline
        194 & Stahl et al. 2022 & \cite{stahl_computer_2022} & ~ & $\bullet$ & ~ & ~ \\ \hline
        195 & Starke et al. 2022 & \cite{starke_explainability_2022} & ~ & $\bullet$ & ~ & ~ \\ \hline
        196 & Storey et al. 2022 & \cite{storey_explainable_2022} & ~ & ~ & $\bullet$ & ~ \\ \hline
        197 & Strobel and Shokri 2022 & \cite{strobel_data_2022} & $\bullet$ & $\circ$ & $\circ$ & $\circ$ \\ \hline
        198  & Sun et al. 2021 & \cite{Sun.2021} & $\bullet$ & ~ & ~ & $\bullet$ \\ \hline
        199 & Sun et al. 2022 & \cite{sun_investigating_2022} & ~ & ~ & $\bullet$ & ~ \\ \hline
        200  & Suresh et al. 2021 & \cite{Suresh.2021} & $\bullet$ & ~ & $\bullet$ & ~ \\ \hline
        201  & Suriyakumar et al. 2021 & \cite{Suriyakumar.2021} & $\circ$ & ~ & ~ & $\bullet$ \\ \hline
        202 & Svetlova et al. 2021 & \cite{svetlova_ai_2022} & ~ & $\bullet$ & ~ & ~ \\ \hline
        203  & Tan et al. 2022 & \cite{tan_towards_2022} & ~ & ~ & ~ & $\bullet$ \\ \hline
        204  & Tartaglione and Grassetto 2020 & \cite{Tartaglione.2020} & $\bullet$ & $\bullet$ & ~ & $\circ$ \\ \hline
        205 & Terziyan \& Vitko 2022 & \cite{terziyan_explainable_2022} & ~ & ~ & $\bullet$ & ~ \\ \hline
        206 & Thuraisingham 2022 & \cite{thuraisingham_trustworthy_2022} & $\bullet$ & ~ & ~ & ~ \\ \hline
        207 & Tolmejer et al. 2022 & \cite{tolmeijer_capable_2022} & ~ & $\bullet$ & ~ & ~ \\ \hline
        208  & Toreini et al. 2020 & \cite{Toreini.2020} & $\bullet$ & $\bullet$ & $\bullet$ & $\bullet$ \\ \hline
        209 & Tian 2022 & \cite{tian_sphinx_2022} & ~ & ~ & ~ & $\bullet$ \\ \hline
        210 & Tiddi and Schlobach 2022 & \cite{tiddi_knowledge_2022} & ~ & ~ & $\bullet$ & ~ \\ \hline
        211  & Tran et al. 2021 & \cite{AnhTuTran.2021} & ~ & ~ & ~ & $\bullet$ \\ \hline
        212 & Tsamados et al. 2022 & \cite{tsamados_ethics_2022} & ~ & $\bullet$ & ~ & ~ \\ \hline
        213 & Tsiakis and Murray 2022 & \cite{tsiakas_using_2022} & ~ & ~ & $\bullet$ & ~ \\ \hline
        214 & Utomo et al. 2022 & \cite{utomo_federated_2022} & $\circ$ & ~ & ~ & $\bullet$ \\ \hline
        215 & Valentine 2022 & \cite{valentine_recommender_2022} & ~ & $\bullet$ & ~ & ~ \\ \hline
        216 & Vakkuri 2021 & \cite{VilleVakkuri.2021} & $\circ$ & $\bullet$ & ~ & ~ \\ \hline
        217 & Vakkuri et al. 2022 & \cite{vakkuri_how_2022} & ~ & $\bullet$ & ~ & ~ \\ \hline
        218 & Vellido et al. 2020 & \cite{Vellido.2020} & ~ & ~ & $\bullet$ & ~ \\ \hline
        219 & Vilone and Logo 2021 & \cite{GiuliaVilone.2021} & $\bullet$ & ~ & $\bullet$ & ~ \\ \hline
        220 & Waller and Waller 2022 & \cite{waller_assembled_2022} & ~ & $\bullet$ & ~ & ~ \\ \hline
        221 & Wang et al. 2020 & \cite{Wang.2020} & ~ & ~ & ~ & $\bullet$ \\ \hline
        222 & Wang et al. 2022 & \cite{wang_flare_2022} & ~ & ~ & ~ & $\bullet$ \\ \hline
        223 & Wang et al. 2022 & \cite{wang_blockchain_2022} & ~ & ~ & ~ & $\bullet$ \\ \hline

    \end{tabular}
    \caption{Qualitative Analysis 7/8}
    \label{tab:qualitytive_analysis_7_8}
\end{table}

\begin{table}
    \rowcolors{2}{gray!25}{white}
    \begin{tabular}{|l|l|l|l|l|l|l|}
        \hline
        \rowcolor{gray!50}
        ~ & Author & Ref. & Tr. AI & Eth. AI & XAI & PP \& Sec. AI \\ \hline

        224 & Wang et al. 2022 & \cite{wang_revise_2022} & ~ & $\bullet$ & ~ & ~ \\ \hline
        225 & Wang and Moulden 2021 & \cite{Wang.2021} & $\bullet$ & ~ & ~ & ~ \\ \hline
        226 & Watson 2022 & \cite{watson_agree_2022} & ~ & ~ & $\bullet$ & ~ \\ \hline
        227 & Weinberg 2022 & \cite{weinberg_rethinking_2022} & ~ & $\bullet$ & ~ & ~ \\ \hline
        228 & Werder et al. 2022 & \cite{werder_establishing_2022} & $\circ$ & $\bullet$ & $\circ$ & $\circ$ \\ \hline
        229 & Wibawa 2022 & \cite{wibawa_homomorphic_2022} & ~ & ~ & ~ & $\bullet$ \\ \hline
        230 & Wing 2021 & \cite{Wing.2021} & $\bullet$ & $\circ$ & $\bullet$ & $\circ$ \\ \hline
        231 & Wyhmeister et al. 2022 & \cite{vyhmeister_responsible_2022} & ~ & $\bullet$ & ~ & ~ \\ \hline
        232 & Xiaoling et al. 2021 & \cite{XianxianLi.2021} & $\bullet$ & $\bullet$ & ~ & ~ \\ \hline
        233 & Xu et al. 2021 & \cite{Xu.2021} & ~ & ~ & ~ & $\bullet$ \\ \hline
        234 & Xu et al. 2021 & \cite{Xu_Xhu.2021} & ~ & ~ & ~ & $\bullet$ \\ \hline
        235 & Yang et al. 2021 & \cite{Yang.2021} & ~ & ~ & ~ & $\bullet$ \\ \hline
        236 & Yang et al. 2022 & \cite{yang_unbox_2022} & ~ & $\bullet$ & ~ & ~ \\ \hline
        237 & Yang et al. 2022 & \cite{yang_trustworthy_2022} & ~ & ~ & ~ & $\bullet$ \\ \hline
        238 & Yuan and Shen 2020 & \cite{Yuan.2020} & ~ & ~ & ~ & $\bullet$ \\ \hline
        239 & Yuan et al. 2020 & \cite{Yuan.2020b} & $\circ$ & ~ & $\bullet$ & ~ \\ \hline
        240 & Zapechnikov et al. & \cite{SergeyZapechnikov.2020} & ~ & ~ & ~ & $\bullet$ \\ \hline
        241 & Zhang et al. 2021 & \cite{Zhang.2021} & $\bullet$ & $\bullet$ & ~ & $\circ$ \\ \hline
        242 & Zhang et al. 2021 & \cite{Zhang.Qin.Li.2021} & $\bullet$ & $\circ$ & $\circ$ & $\circ$ \\ \hline
        243 & Zhang et al. 2020 & \cite{Zhang.Yiu.Hui.2020} & ~ & ~ & ~ & $\bullet$ \\ \hline
        244 & Zhang et al. 2022 & \cite{zhang_debiased-cam_2022} & ~ & $\circ$ & $\bullet$ & ~ \\ \hline
        245 & Zhang et al. 2022 & \cite{zhang_visual_2022} & ~ & ~ & ~ & $\bullet$ \\ \hline
        246 & Zhao et al. 2022 & \cite{zhao_pvd-fl_2022} & ~ & ~ & ~ & $\bullet$ \\ \hline
        247 & Zhao et al. 2022 & \cite{zhao_correlated_2022} & ~ & ~ & ~ & $\bullet$ \\ \hline
        248 & Zhou et al. 2020 & \cite{Zhou.2020} & $\bullet$ & $\bullet$ & ~ & $\circ$ \\ \hline
        249 & Zhou et al. 2020 & \cite{Zhou.2020b} & ~ & ~ & ~ & $\bullet$ \\ \hline
        250 & Zhou et al. 2022 & \cite{zhou_human---loop-aided_2020} & ~ & ~ & ~ & $\bullet$ \\ \hline
        251 & Zhou et al. 2022 & \cite{holzinger_towards_2022} & ~ & $\bullet$ & ~ & ~ \\ \hline
        252 & Zhu et al. 2020 & \cite{Zhu.2020} & ~ & ~ & ~ & $\bullet$ \\ \hline
        253 & Zhu et al. 2022 & \cite{zhu_fednkd_2022} & ~ & ~ & ~ & $\bullet$ \\ \hline
        254 & Zytek et al. 2021 & \cite{Zytek.2021} & ~ $\bullet$ & ~ & $\bullet$ & ~ \\ \hline

    \end{tabular}
    \caption{Qualitative Analysis 8/8}
    \label{tab:qualitytive_analysis_8_8}
\end{table}

\newpage

\section{Discussion}
\label{section:discussion}
Several key points have emerged from the analysis. It has become clear that AI will have an ever-increasing impact on our daily lives, from delivery robots to e-health, smart nutrition and digital assistants, and the list is growing every day. AI should be viewed as a tool, not a system that has infinite control over everything. It should therefore not replace humans or make them useless, nor should it lead to humans no longer using their own intelligence and only letting AI decide. We need a system that we can truly call "responsible" AI.
The analysis has clearly shown that the elements of ethics, privacy, security and explainability are the true pillars of responsible AI, which should lead to a basis of trust.\\

\subsection{Pillars of Responsible AI }

Here we highlight the most important criteria that a responsible AI should fulfill. These are also the points that a developer should consider if he wants to develop responsible AI. Therefore, they also form the pillars for the future framework.\\

Key-requirements for the Ethical AI are as follows:
\begin{itemize}
    \item fair: non-biased and non-discriminating in every way,
    \item accountability: justifying the decisions and actions,
    \item sustainable: built with long-term consequences in mind, satisfying the Sustainable Development Goals,
    \item compliant: with robust laws and regulations.

\end{itemize}

Key-requirements for the privacy and security techniques are identified as follows:
\begin{itemize}
    \item need to comply with regulations: HIPAA, COPPA, and more recently the GDPR (like, for example, the Federated Learning),
    \item need to be complemented by proper organizational processes,
    \item must be used depending on tasks to be executed on the data and on specific transactions a user is executing,
    \item use hybrid PPML-approaches because they can take advantage of each component, providing an optimal trade-off between ML task performance and privacy overhead,
    \item use techniques that reduce communication and computational cost (especially in distributed approaches).
\end{itemize}

Key-requirements for Explainable AI are the following:
\begin{itemize}
    \item Human-Centered: the user interaction plays a important role and how he understands and interacts with the system,
    \item Explanations must be tailored to the user needs and target group
    \item Intuitive User interface/experience: the results need to be presented in a understandable visual language,
    \item Explainable is also feature to say how well the system does its work (non functional requirement),
    \item Impact of explanations on decision making process,
\end{itemize}

Key-Perceptions of trustworthy AI are as follows:

\begin{itemize}
    \item ensure user data is protected,
    \item probabilistic accuracy under uncertainty,
    \item provides an understandable, transparent, explainable reasoning process to the user,
    \item usability,
    \item act "as intended" when facing a given problem,
    \item perception as fair and useful,
    \item reliability.
\end{itemize}

We define Responsible AI as an interdisciplinary and dynamic process:
it goes beyond technology and includes laws (compliance and regulations) and society standards such as ethics guidelines and the Sustainable Development Goals.\\

 \begin{figure}[h]
     \centering
     \includegraphics[height=8cm]{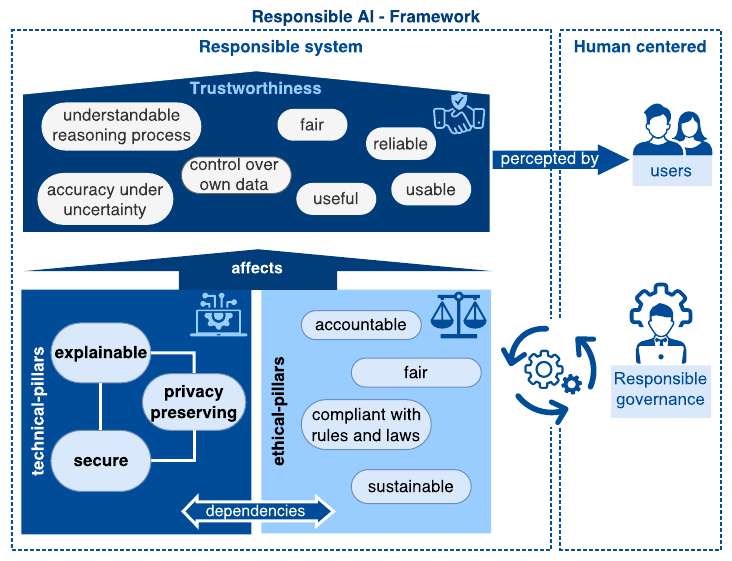}
     \caption{Pillars of the Responsible AI framework}
     \label{fig:framework}
 \end{figure}

Figure \ref{fig:framework} shows that on the one hand there are social/ethical requirements/pillars and on the other hand the technical requirements/pillars. All of them are dependent on each other. If the technical and ethical side is satisfied the user trust is maintained. Trust can be seen as the perception of the users of AI.\\
There are also "sub-modules" present in each of the pillars, like accountability, fairness, sustainability and compliance in the field of ethics. They are crucial that we can say the AI meets ethical requirements. \\
Furthermore, the explainability methods must value privacy, meaning they must not have that much access to a model so that it results in a privacy breach. Privacy is dependent on security, because security is a prerequisite for it.\\
With each  "responsible system" there are the humans that care for the system. The people who take care of the system must also handle it responsibly and constantly carry out maintenance work and check by metrics whether the responsibility is fulfilled. This can be ensured by special metrics which are considered as a kind of continuous check as standard. This means responsible AI encompasses the system-side and the developer-side.\\
Human-Centered AI (mentioned in \ref{section:aspects_of_responsible_ai}) needs to be considered as a very important part of responsible AI and it is closely connected to the approach "Human-in-the-loop". The human in the loop here is very important because this is the person who checks and improves the system during the life cycle. so the whole responsible AI system needs to be Human-Centered, too. This topic will not be dealt with in detail in this study, but is a part of the future work.\\
Therefore, responsible AI is interdisciplinary, and it is not a static but it is a dynamic process that needs to be taken care of in the whole system lifecycle.

\subsection{Trade-offs}
To fulfill all aspects comes with tradeoffs as discussed for example in \cite{strobel_data_2022} and comes for example at cost of data privacy. For example the methods that make model more robust against attacks or methots that try to explain a models behaviour and could leak some information. But we have fo find a way to manage that AI Systems that are accurate, fair, private, robust and explainable at the same time, which will be a very challenging task. We think that one approach to start with would be to create a benchmark for the different requirements that can determine to which proportion a certain requirement is fulfilled, or not.

\section{Research Limitations}
In the current study, we have included the literature available through various journals and provided a comprehensive and detailed survey on the literature in the field of responsible AI.\\
In conducting the study, we unfortunately had the limitation that some journals were not freely accessible despite a comprehensive access provided by our institutions. Although we made a good effort to obtain the information needed for the study on responsible AI from various international journals, accessibility was still a problem. It is also possible that some of the relevant research publications are not listed in the databases we used for searching. Additional limitation is the time frame of searched articles; this was carefully addressed to include only the state-of-the-art in the field. However, some older yet still current developement might have been missed out.

\section{Conclusion}
The field of AI is such a fast changing area and a legal framework for responsible AI is strongly necessary.
From the series of EU-Papers on Artificial Intelligence of the last 2 years we noticed that "trustworthy AI" and "responsible AI" are not clearly defined, and as such a legal framework could not be efficiently established.
Hence, the trust as a goal to define a framework/regulation for AI is not sufficient. Regulations for 'responsible AI' need to be defined instead. As the EU is a leading authority when it comes to setting standards (like the GDPR) we find it is absolutely necessary to help the politicians to really know what they are talking about. On the other hand, helping practitioners to prepare for what is coming next in both research and legal regulations is also of great importance.\\
The present research made important contributions to the concept of responsible AI. It is the first contribution to wholly address the "responsible AI" by conducting a structured literature research, and an overarching definition is presented as a result. The structured literature review covered 118 most recent high quality works on the  topic. We have included a qualitative and quantitative analysis of the papers covered.\\
By defining "responsible AI" and further analyzing the state of the art of its components (i.e., Human-centered, Trustworthy, Ethical, Explainable, Privacy(-preserving) and Secure AI), we have shown which are the most important parts to consider when developing AI products and setting up legal frameworks to regulate their development and use.\\
In the discussion section we  have outlined an idea for developing a future framework in the context of Responsible AI based on the knowledge and insights gained in the analysis part.\\
In future research the topic of Human-Centered AI and "Human-in-the-loop" should be developed further in the context responsible AI. Other important topics to be worked upon are the benchmarking approaches for responsible AI and a holistic framework for Responsible AI as the overarching goal.

\bibliographystyle{vancouver}
\bibliography{literature}

\end{document}